\documentclass[letterpaper]{article} % DO NOT CHANGE THIS
\usepackage{aaai2026}  % DO NOT CHANGE THIS
\usepackage{times}  % DO NOT CHANGE THIS
\usepackage{helvet}  % DO NOT CHANGE THIS
\usepackage{courier}  % DO NOT CHANGE THIS
\usepackage[hyphens]{url}  % DO NOT CHANGE THIS
\usepackage{graphicx} % DO NOT CHANGE THIS
\usepackage{natbib}  % DO NOT CHANGE THIS AND DO NOT ADD ANY OPTIONS TO IT
\usepackage{caption} % DO NOT CHANGE THIS AND DO NOT ADD ANY OPTIONS TO IT
\usepackage{algorithm}
\usepackage{algorithmic}

\usepackage{newfloat}
\usepackage{listings}
\DeclareCaptionStyle{ruled}{labelfont=normalfont,labelsep=colon,strut=off} % DO NOT CHANGE THIS
\floatstyle{ruled}
\newfloat{listing}{tb}{lst}{}
\floatname{listing}{Listing}
\usepackage[dvipsnames]{xcolor}

\definecolor{Black}{rgb}{0.0, 0.0, 0.0}

\definecolor{DarkGreen}{rgb}{0.10, 0.55, 0.10} % ✔️
\definecolor{DeepSkyBlue3}{rgb}{0.0, 0.686, 0.843}
\definecolor{DarkTurquoise}{rgb}{0.0, 0.843, 0.843}
\definecolor{Cyan3}{rgb}{0.0, 0.843, 0.686}
\definecolor{LightSeaGreen}{rgb}{0.0, 0.686, 0.686} % ✔️

\definecolor{RoyalBlue}{rgb}{0.20, 0.60, 0.86}
\definecolor{DeepSkyBlue}{rgb}{0.0, 0.686, 1.}

\definecolor{DodgerBlue}{rgb}{0.0, 0.529, 1.} % ✔️
\definecolor{DodgerBlue2}{rgb}{0.0, 0.3725, 1.}
\definecolor{DodgerBlue3}{rgb}{0.0, 0.3725, 0.843}
\definecolor{DarkCyan}{rgb}{0.0, 0.54, 0.54}

\definecolor{Gray}{gray}{0.9}
\definecolor{ChromeYellow}{rgb}{1.0, 0.65, 0.0}

\definecolor{Gold}{rgb}{1.0, 0.843, 0.0}

\definecolor{Crimson}{rgb}{0.86, 0.08, 0.24}
\definecolor{IndianRed}{rgb}{1.0, 0.373, 0.529}

\definecolor{SunsetOrange}{rgb}{0.98, 0.37, 0.33}
\definecolor{DarkOrange}{rgb}{1.0, 0.529, 0.}
\ifx \submission \undefined

\else
\fi

\newcommand{\ie}{\textit{i.e.}}
\newcommand{\eg}{\textit{e.g.}}

\usepackage{amsmath}
\usepackage{amssymb}
\usepackage{booktabs}
\usepackage[table]{xcolor}
\usepackage{multirow}
\usepackage{array}
\newcolumntype{C}[1]{>{\centering\arraybackslash}p{#1}} % fixed-width centered
\usepackage{pifont}
\title{Alfa: Attentive Low-Rank Filter Adaptation\\for Structure-Aware Cross-Domain Personalized Gaze Estimation}
\author{
    He-Yen Hsieh, Wei-Te Mark Ting, H.T. Kung
}
\affiliations{
    Harvard University

}

\usepackage{bibentry}
\begin{document}

\maketitle

\begin{abstract}
Pre-trained gaze models learn to identify useful patterns commonly found across users, but subtle user-specific variations (\ie, eyelid shape or facial structure) can degrade model performance.
Test-time personalization (TTP) adapts pre-trained models to these user-specific domain shifts using only a few unlabeled samples.
Efficient fine-tuning is critical in performing this domain adaptation: data and computation resources can be limited---especially for on-device customization.
While popular parameter-efficient fine-tuning (PEFT) methods address adaptation costs by updating only a small set of weights, they may not be taking full advantage of structures encoded in pre-trained filters.
To more effectively leverage existing structures learned during pre-training, we reframe personalization as a process to reweight existing features rather than learning entirely new ones.

We present Attentive Low-Rank Filter Adaptation (Alfa) to adapt gaze models by reweighting semantic patterns in pre-trained filters.
With Alfa, singular value decomposition (SVD) extracts dominant spatial components that capture eye and facial characteristics across users.
Via an attention mechanism, we need only a few unlabeled samples to adjust and reweight pre-trained structures, selectively amplifying those relevant to a target user.
Alfa achieves the lowest average gaze errors across four cross-dataset gaze benchmarks, outperforming existing TTP methods and low-rank adaptation (LoRA)-based variants.
We also show that Alfa's attentive low-rank methods can be applied to applications beyond vision, such as diffusion-based language models.
\end{abstract}

\begin{figure}[t!]
\centering
\includegraphics[width=0.75\columnwidth]{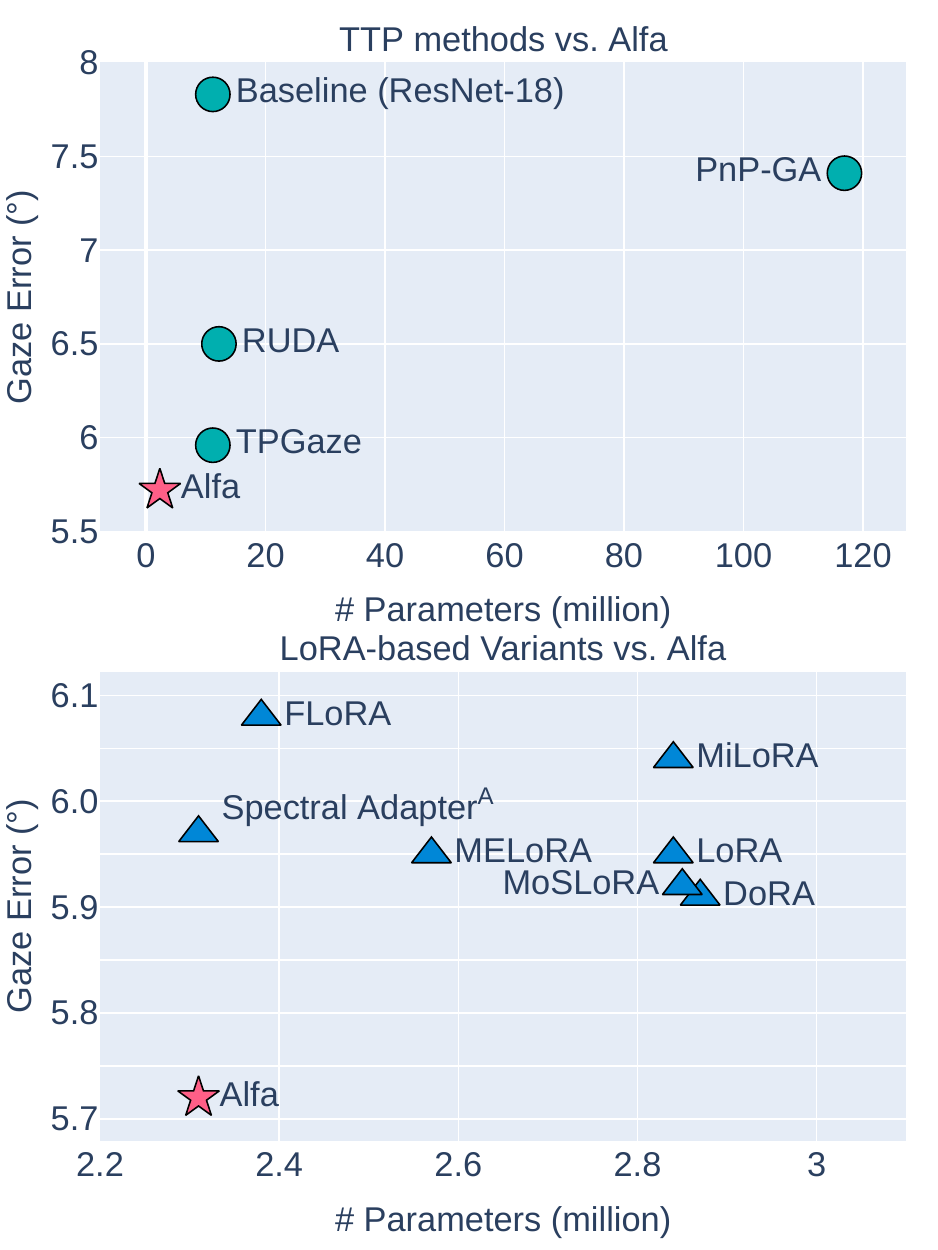} 
\caption{
Alfa achieves the lowest average gaze error, with the smallest model size, across four cross-dataset benchmarks: from ETH-XGaze to MPIIGaze, ETH-XGaze to EyeDiap, Gaze360 to MPIIGaze, and Gaze360 to EyeDiap.
Top: Comparison with other test-time personalization (TTP) methods.
Baseline refers to a ResNet-18 without fine-tuning.
Bottom: Comparison with low-rank adaptation (LoRA)-based variants.
}
\label{fig:teaser}
\end{figure}

\section{Introduction}

Gaze estimation can infer the direction a person is looking from facial or eye-region images.
This capability is central to many applications in augmented reality, human-computer interaction, and assistive technologies.
For example, in gaze-assisted communication systems \cite{LeeWBCRF24, KhanN0V22}, detecting the user's eye focus is essential for delivering intuitive and effective responses.

Gaze estimation models perform well under controlled conditions, but may struggle in real-world settings.
Differences across users, camera configurations, and environments, such as changes in lighting or head pose, create discrepancies between training and deployment conditions. 
These domain shifts decrease accuracy when models lack robustness outside their initial training conditions.

\begin{figure*}[t!]
\centering
\includegraphics[width=0.7\textwidth]{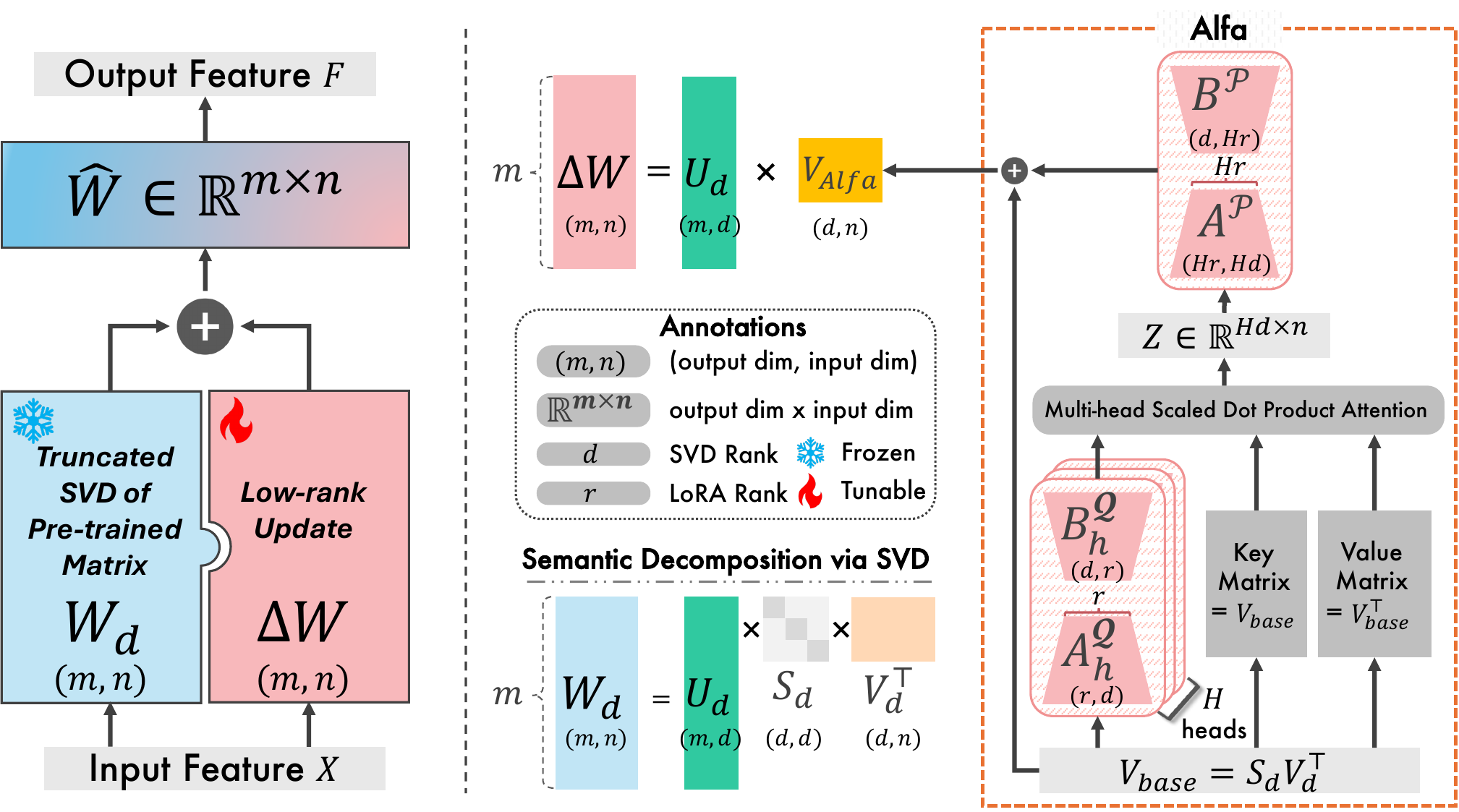}
\caption{
Overview of Attentive Low-Rank Filter Adaptation (Alfa).
(a) The pre-trained weight matrix is approximated using truncated SVD: $W_d = U_d S_d V_d^\top$.
Then, a tunable low-rank update $\Delta W$ is added for adaptation.
(b) Alfa adapts gaze models by reweighting spatial structures encoded in pre-trained filters.
Alfa extracts dominant spatial patterns ($V_{\text{base}} = S_d V_d^\top$) via singular value decomposition (SVD).
For personalization, multi-head low-rank modules $A^\mathcal{Q}$ and $B^\mathcal{Q}$ generate query weights, and $V_{\text{base}}$ and $V_{\text{base}}^\top$ are reused as key and value matrices. 
Using multi-head scaled dot-product attention, Alfa identifies the spatial structures most relevant to a target user.
Alfa aggregates this into a personalized update using additional low-rank modules $A^\mathcal{P}$ and $B^\mathcal{P}$, forming $V_{\text{Alfa}}$, which encodes gaze-specific adaptations informed by the pre-trained spatial structure.
}
\label{fig:alfa}
\end{figure*}

Test-time personalization (TTP) \cite{LiuQLHWPY24, BaoLWL22, LiuLWL21} is a variant of unsupervised domain adaptation (UDA) \cite{WangJLNDLXL22, BaoLWL22, LiuLWL21} in which the model adapts to a new user during deployment, relying only on unlabeled samples collected at test time. 
TTP provides privacy-preserving, on-device adaptation in a few-shot setting and offers a practical solution for real-world deployment without access to the original training data or ground-truth gaze labels.
TTP also addresses some key challenges in gaze estimation: differences in users' eye shapes, appearances, or camera placement may degrade model performance.

Intuitively, human faces may exhibit spatially coherent patterns (\ie, eye and facial geometry) \cite{TianTHWZ23, Hong2022}.
With sufficient pre-training, models learn to capture useful general features that correspond to these patterns, encoding them into spatial structures (\eg~filters) within the weights.
Yet, small changes to these patterns can significantly reduce prediction accuracy.
While fine-tuning can help mitigate this, leveraging more information from the pre-trained model may improve performance \cite{WangLWCC25, MengWZ24}.
We find that leveraging existing features rather than learning entirely \textit{new} features is effective for data-scarce adaptation in gaze estimation (\eg, with only five new images).

In this paper, we propose Attentive Low-Rank Filter Adaptation (Alfa), which adapts gaze models by reweighting the \textit{spatial structure} encoded in pre-trained filters. 
As depicted in Figure \ref{fig:alfa}, Alfa first performs SVD on pre-trained weights to obtain dominant spatial structures that correspond to common patterns (\textit{e.g.}, geometric features in the eye and face).
Instead of learning new filters or initializing new weights, Alfa adapts models by modulating the influence of these dominant patterns by using a few unlabeled samples from a target user.
This enables personalized predictions from minimal additional data while maintaining alignment with the original learned representation.
Visualization results demonstrate that Alfa attends to common and intuitive localized regions, such as the eyelids, which vary across users and should be effectively detected during adaptation (See Figure \ref{fig:lora_vs_alfa}).

In summary, our contributions are as follows:
\begin{enumerate}
    \item Attentive Low-Rank Filter Adaptation (Alfa) adapts gaze models by attending over structured spatial patterns derived from SVD, instead of treating weights as unstructured tensors.
    \item Within Alfa, a multi-head low-rank adaptation module accepts scalable personalization capacity during fine-tuning. Storing the pre-trained weight in truncated SVD form reduces model size, and makes Alfa's updates fully mergeable without increasing model size at deployment (See Section \ref{sec:trn_tst}).
    \item Empirical demonstration of Alfa outperforming prior methods on four cross-domain gaze benchmarks using only a few unlabeled test-time samples.
    \item Extension of Alfa's structured adaptation to diffusion-based large language models (LLMs), showing improved zero-shot reasoning across multiple benchmarks.
\end{enumerate}

\section{Related Work}
\subsection{Test-Time Personalization (TTP)}
To improve generalization across domains, many gaze estimation methods adopt unsupervised domain adaptation (UDA). 
In the standard UDA setting \cite{LiuLWL21, GuoYZCLZ20, KellnhoferRSM019}, models are trained with labeled source-domain data and unlabeled target-domain data. 
PnP-GA \cite{LiuLWL21} uses an outlier-guided loss to improve adaptation with a few source samples and unlabeled target data.
To relax the need for source data, source-free UDA methods \cite{BaoLWL22, WangJLNDLXL22} adapt models using only unlabeled target data, though they still require a relatively large amount of it.
RUDA~\cite{BaoLWL22}, for example, leverages rotation-based gaze synthesis and geometric constraints to improve pseudo-label quality.
Personalization provides a complementary way by adapting models to individual users. 
While few-shot methods \cite{GhoshHDK22, ChenS20, ParkMMIHK19, YuLO19} fine-tune models with a small number of labeled personal samples, obtaining gaze labels is often expensive or impractical. 
TTP \cite{LiuQLHWPY24} addresses this challenge by adapting the model to each user during inference using a few unlabeled personal samples.
TTP is well-suited for on-device scenarios, where user appearance varies and updating the full model is often too resource-intensive.
These constraints motivate efficient, lightweight adaptation methods that support per-user personalization without requiring labeled data or access to the source domain.
Alfa supports test-time personalization with no additional inference cost, is scalable through multi-head adaptation during fine-tuning, and uses SVD to select critical semantic components, resulting in a compact and efficient model.

\subsection{Low-Rank Adaptation}
\label{sec:related_work_lora}
Low-rank adaptation injects small trainable matrices into pre-trained models to fine-tune them efficiently without altering the original weights.
LoRA \cite{HuSWALWWC22} is a widely adopted method that adds rank-constrained updates to existing layers and has shown strong performance across various tasks.
Building on this idea, MiLoRA \cite{WangLWCC25} initializes adaptation weights using principal components from SVD, while PiSSA \cite{MengWZ24} directly updates the top spectral components.
Spectral Adapter$^A$ \cite{ZhangP24} introduces learnable weights applied to the spectral bases, and MoSLoRA \cite{WuW0W24} places a mixer matrix between LoRA modules to improve representation capacity.
DoRA \cite{LiuWY0WCC24} decomposes pre-trained weights into independent direction and magnitude components.
MELoRA \cite{RenS0ZRRCP24} ensembles multiple mini-expert LoRA branches, and FLoRA \cite{SiW0XLDQ0025} applies Tucker decomposition to capture multi-dimensional parameter changes through a shared low-rank core.
Despite their varied adaptation strategies, they often treat model weights as unstructured tensors and overlook the spatial structure encoded in pre-trained filters.
In contrast, Alfa attends to these spatial structures and reweights them during personalization, enabling structure-aware adaptation guided by meaningful semantics.

\section{Alfa}

\subsection{Problem Definition and Preliminary}
We consider the TTP task for gaze estimation, formulated as a variant of UDA.
The goal is to adapt a gaze model trained on a general source domain $\mathcal{D}_S$ to a new, unseen user in a target domain $\mathcal{D}_T$, with only a few unlabeled samples.
Let each RGB input image be denoted as $I$, and its ground-truth gaze direction represented by a 2D vector $g \in \mathbb{R}^2$ (yaw and pitch).
The source domain provides labeled data $\mathcal{D}_S = \{(I_i^S, g_i^S)\}_{i=1}^{N_S}$, while the target domain provides a small unlabeled set $\mathcal{D}_T = \{ I_k^T \}_{k=1}^{N_T}$, typically with $N_T = 5$.
In a gaze model, we denote the input and output features of a convolutional or linear layer as $X \in \mathbb{R}^{n \times h \times w}$ and $F \in \mathbb{R}^{m \times h' \times w'}$, respectively, where $n$ and $m$ are the input and output channel dimensions, and $h, w$ and $h', w'$ are the height and width of the input and output spatial resolutions.
These features are connected through a pre-trained weight matrix $W \in \mathbb{R}^{m \times n}$.

To enable data- and parameter-efficient adaptation from a few unlabeled samples, we build on LoRA, which injects a trainable low-rank update into a pre-trained weight matrix: $\Delta W = A B$, where $A \in \mathbb{R}^{m \times r}$, $B \in \mathbb{R}^{r \times n}$, and $r \ll \min(m, n)$, with \( r \) as the LoRA rank.
Typical initializations set $A \sim \mathcal{N}(0, \sigma^2)$ and $B = 0$.
However, this formulation overlooks the spatial and geometric structure embedded in the pre-trained weights: structure that captures critical visual patterns relevant to gaze across users.
Alfa addresses this limitation by personalizing gaze models through structured reweighting of semantic filters extracted from pre-trained weights.
Section \ref{sec:svd} introduces SVD-based decomposition of pre-trained weights to extract a semantic basis dictionary.
Section \ref{sec:alfa} describes the structured reweighting mechanism that selectively adapts semantic components for personalization.

\subsection{Structured Decomposition of Gaze Filters}
\label{sec:svd}
Gaze estimation relies on structured visual patterns shaped by the anatomy of the eyes and surrounding facial regions.
While these patterns vary across individuals, the variations tend to follow a few set of consistent spatial changes, such as shifts in iris position or subtle deformations in surrounding facial muscles.
These consistencies suggest that pre-trained weights encode recurring spatial structures that are broadly shared across users (see Section \ref{sec:vis_pretrain}). 

When trained on a diverse population, the model learns to encode this structure in its weights, forming a strong foundation for general gaze prediction.
To make this structure explicit, we apply SVD to the pre-trained weight matrix. 
Let \( W \in \mathbb{R}^{m \times n} \) denote the weight matrix of a convolutional or linear layer.
We compute a truncated SVD:
\begin{equation}
W \approx W_d = U_d S_d V_d^\top
\end{equation}
where \( d \ll \min(m, n) \) is the target rank, yielding:
\begin{itemize}
    \item \( U_d \in \mathbb{R}^{m \times d} \): output projection matrix (left singular vectors),
    \item \( S_d \in \mathbb{R}^{d \times d} \): singular values representing the importance of each direction,
    \item \( V_d^\top \in \mathbb{R}^{d \times n} \): dominant spatial directions in input space.
\end{itemize}

\noindent
Thus, we obtain the \textbf{semantic basis dictionary}, defined as:
\[
V_{\text{base}} = S_d V_d^\top \in \mathbb{R}^{d \times n} 
\]
$V_{\text{base}}$ reflects the highest-energy components learned during gaze pre-training.
These components capture key spatial patterns relevant to gaze behavior (\eg, iris and peri-ocular cues co-activating).
Components reused most frequently during pre-training to reduce gaze loss have the highest energy.
Since SVD ranks weight-space patterns by energy, truncating to these leading components preserves the dominant gaze-relevant structure, yielding a compact, structure-aware basis for personalization.
Retaining only the top \( d \) components ensures we focus on the most expressive, information-rich parts of the filter space while providing a compact basis for downstream adaptation.
In Alfa, this semantic structure facilitates learning efficient, personalized updates in later stages.

\subsection{Personalizing the Semantic Basis Dictionary}
\label{sec:alfa}
The semantic basis dictionary extracted via SVD provides a compact set of spatial patterns that generalize well across individuals in gaze estimation.
However, these patterns may not fully capture the unique appearance characteristics of each user.
Alfa introduces a low-rank update that reweights the components of the semantic basis dictionary without discarding the shared structure learned during pre-training.
This approach allows the model to adapt to individual differences while preserving gaze-relevant information encoded in the pre-trained filters.
To personalize the model, we add a low-rank update $\Delta W$ on top of the pre-trained weight \( W_d \). The adapted weight is defined as:
\begin{equation}
\hat{W} = W_d + \Delta W
\end{equation}
where \( \Delta W \in \mathbb{R}^{m \times n} \) is a low-rank personalization term computed as $\Delta W = U_d V_{\text{alfa}}$, 
where, \( V_{\text{alfa}} \) is a learnable update produced by the Alfa module.

\paragraph{Attending to Semantic Basis Dictionary}
Alfa computes a personalized adaptation by applying a multi-head attention mechanism over the semantic basis dictionary $V_{\text{base}} \in \mathbb{R}^{d \times n}$.
The dictionary captures shared spatial patterns from pre-training, and Alfa uses attention to reweight the slices most relevant to the target subject.
Let $H$ be the number of attention heads.
Different heads attend to different rank slices, combining complementary cues and reducing drift when a few personal samples are available (see supplementary material Section F).
For each attention head indexed by $h \in \{1, \dots, H\}$, we define a pair of low-rank projection matrices: $A^{\mathcal{Q}}_h \in \mathbb{R}^{r \times d}$ for the query projection, and $B^{\mathcal{Q}}_h \in \mathbb{R}^{d \times r}$ for the query back-projection.
The query projections are initialized as
\begin{equation}
A^{\mathcal{Q}}_h \sim \mathcal{N}(0, \sigma^2), \quad B^{\mathcal{Q}}_h = 0,
\end{equation}
where $\sigma$ denotes the standard deviation of the initialization distribution.
Each head computes a query matrix as
\begin{equation}
\mathcal{Q}_h = B^{\mathcal{Q}}_h A^{\mathcal{Q}}_h V_{\text{base}} \in \mathbb{R}^{d \times n}
\end{equation}
Key and value matrices are directly derived from $V_{\text{base}}$ and shared across all attention heads.
The key matrix is defined as $\mathcal{K} = V_{\text{base}} \in \mathbb{R}^{d \times n}$.
The value matrix is its transpose, $\mathcal{V} = V_{\text{base}}^\top \in \mathbb{R}^{n \times d}$.
For each head, scaled dot-product attention is computed using its query $\mathcal{Q}_h$ as:
\begin{equation}
\text{Attn}_h = \text{softmax}\left( \frac{\mathcal{Q}_h \mathcal{K}^\top}{\sqrt{n}} \right) \in \mathbb{R}^{d \times d}
\end{equation}
\begin{equation}
Z_h = \mathcal{V} \, \text{Attn}_h^\top \in \mathbb{R}^{n \times d}
\end{equation}
Each output $Z_h$ is transposed and then stacked across heads:
\begin{equation}
Z = [Z_1^\top, \dots, Z_H^\top] \in \mathbb{R}^{Hd \times n}
\end{equation}

\paragraph{Integrating Multi-Head Adaptation}
After aggregating the multihead outputs into \( Z \in \mathbb{R}^{Hd \times n} \), we project them back into the semantic space using two low-rank matrices, \( A^{\mathcal{P}} \in \mathbb{R}^{rH \times Hd} \) and \( B^{\mathcal{P}} \in \mathbb{R}^{d \times rH} \).
These are initialized as:
\begin{equation}
A^{\mathcal{P}} \sim \mathcal{N}(0, \sigma^2), \quad B^{\mathcal{P}} = 0
\end{equation}
We compute the personalized update as:
\begin{equation}
V_{\text{Alfa}} = B^{\mathcal{P}} A^{\mathcal{P}} Z + V_{\text{base}} \in \mathbb{R}^{d \times n}
\end{equation}
This completes the adaptation process, resulting in a low-rank update $\Delta W = U_d V_{\text{Alfa}}$.

\begin{figure*}[t!]
\centering
\includegraphics[width=0.9\textwidth]{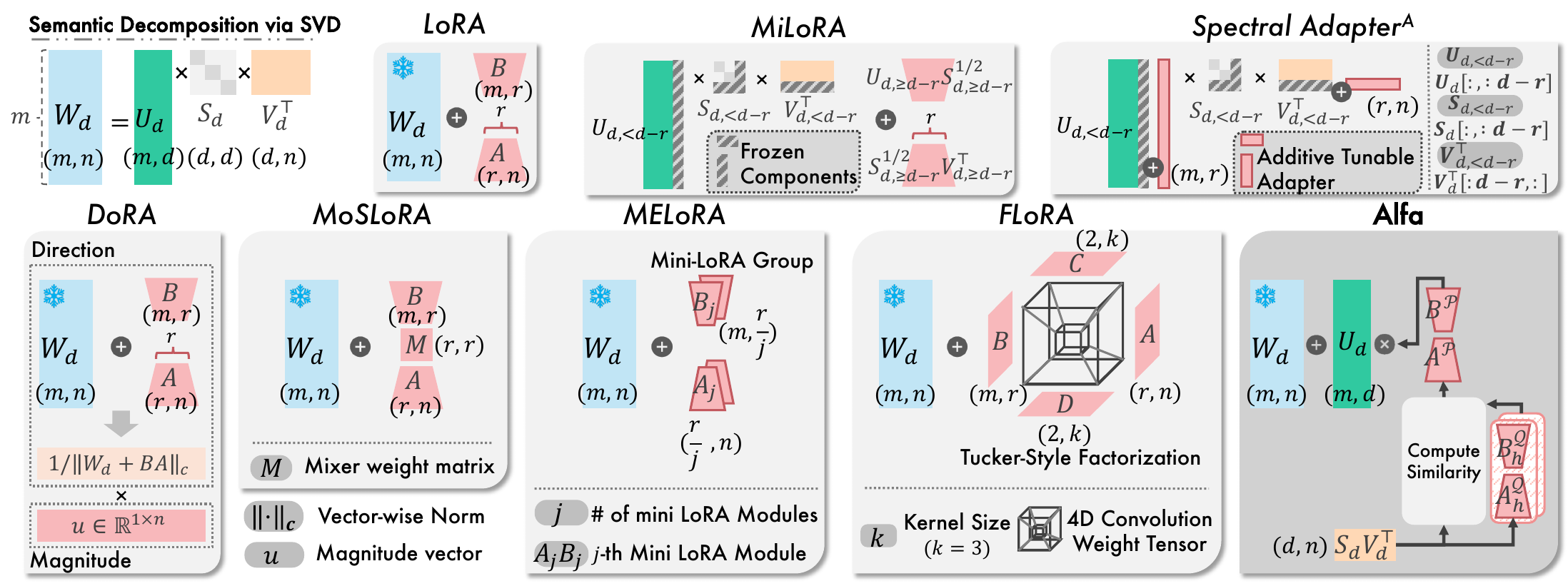}
\caption{
Comparison with other LoRA-based variants, including LoRA, MiLoRA, Spectral Adapter$^A$, DoRA, MoSLoRA, MELoRA, and FLoRA.
Alfa selectively reuses semantic patterns encoded in pre-trained weights and activates the most relevant ones during adaptation. 
Only blocks with red backgrounds are tunable. Best viewed in color.
}
\label{fig:lora_variants}
\end{figure*}

\subsection{Fine-tuning and Inference with Alfa}
\label{sec:trn_tst}

We describe the fine-tuning procedure for Alfa and discuss how it maintains computational efficiency during inference, despite the inclusion of multi-head LoRA modules.
\paragraph{Fine-Tuning} 
Human faces typically exhibit left-right symmetry, and gaze behavior should remain consistent under horizontal flips.
To exploit this property, we apply a symmetry loss following \citet{KellnhoferRSM019} using the five unlabeled personal samples $\{I_k^T\}_{k=1}^{N_T}$, where $N_T = 5$.
Let $f_\theta(\cdot)$ denote the gaze model.
For each image, we generate a horizontally flipped version $I_k^{T,\text{flip}}$ and obtain predictions:
\begin{equation}
\hat{g}_k^T = f_\theta(I_k^T), \quad \hat{g}_k^{T,\text{flip}} = f_\theta(I_k^{T,\text{flip}}),
\end{equation}
where $\hat{g} = [\hat{g}_{\text{yaw}}, \hat{g}_{\text{pitch}}] \in \mathbb{R}^2$.
The symmetry loss is computed as:
\begin{equation}
\mathcal{L}_{\text{fine-tune}} = \frac{1}{N_T} \sum_{k=1}^{N_T} \left| \hat{g}_k^T - \text{FlipYaw}(\hat{g}_k^{T,\text{flip}}) \right|_1
\end{equation}
\paragraph{Inference} Adaptation reuses the left basis \( U_d \) from the pre-trained weight \( W_d \), preserving the original model structure.
This enables efficient update merging at inference time.
Specifically, the full adapted weight becomes:
\begin{align}
\hat{W} = W_d + \Delta W &= U_d V_{\text{base}} + U_d V_{\text{Alfa}} \nonumber \\
                         &= U_d \left( V_{\text{base}} + V_{\text{Alfa}} \right)
\end{align}
We denote the sum in parentheses as a factor \( V_{\text{adapt}} \), yielding:
\begin{equation}
\hat{W} = U_d V_{\text{adapt}}, \quad \text{where} \quad V_{\text{adapt}} = V_{\text{base}} + V_{\text{Alfa}}
\end{equation}
Adapted weights stay in low-rank form, keeping the same structure as the original compressed model.
We can simply update the right-hand side to get $\hat{W} = U_d V_{\text{adapt}}$ without first reconstructing the full weight matrix.
In contrast, standard LoRA variants add the low-rank term $AB$ to the full matrix:
\begin{equation}
\hat{W} = W_d + AB = U_d S_d V_d^\top + AB.
\end{equation}
Merging $AB$ necessitates expanding $W_d$ to its full size in $\mathbb{R}^{m \times n}$ and negates the benefit of parameter compression (\textit{e.g.}, model increases from around 2M to 11M parameters).
Storing $W_d$ in its truncated SVD form $U_d S_d V_d^\top$ would preclude directly merging the low-rank term $AB$.
Alfa updates only the SVD right factor \(S_d V_d^\top\) and keeps \(U_d\) fixed, ensuring the adaptation remains low-rank and directly mergeable.

\section{Experiments}

\begin{table*}[t!]
\centering
\setlength{\tabcolsep}{4pt}
\resizebox{0.72\textwidth}{!}{
\begin{tabular}{c|ccc|cccc|c}
\toprule
\multirow{2}{*}{\textbf{Method}} & 
\multicolumn{3}{c|}{\textbf{\# Parameters (M)}} & 
\multicolumn{4}{c|}{\textbf{Source $\rightarrow$ Target Domain (5-shot)}} & 
\multirow{2}{*}{\textbf{Avg}} \\
\cmidrule{2-8}
&
\textbf{Train} & \textbf{Tuned} & \textbf{Test} & 
\( D_E \rightarrow D_M \) & 
\( D_E \rightarrow D_D \) & 
\( D_G \rightarrow D_M \) & 
\( D_G \rightarrow D_D \) & \\
\midrule
Baseline (ResNet-18) & 11.18 & 0 & 11.18 & 8.02 & 7.30 & 7.79 & 8.19 & 7.83 \\
\hline
PnP-GA$^\dagger$ \cite{LiuLWL21} & 116.9 & 116.9 & 116.9 & 6.91 & 7.18 & 7.36 & 8.17 & 7.41 \\
RUDA$^\dagger$ \cite{BaoLWL22} & 12.20 & 12.20 & 12.20 & 6.86 & 6.84 & 6.96 & 5.32 & 6.50 \\
TPGaze \cite{LiuQLHWPY24} & 11.18 & \textbf{0.13} & 11.18 & 6.30 & 5.89 & 6.62 & \textbf{5.04} & 5.96 \\
\hline
\rowcolor{orange!10}
\textbf{Alfa} & 5.26 & 2.98 & \textbf{2.31} & \textbf{5.30} & \textbf{5.82} & \textbf{5.91} & 5.86 & \textbf{5.72} \\
\bottomrule
\end{tabular}
}
\caption{
Comparison with state-of-the-art TTP methods across four cross-domain benchmarks.
Datasets are denoted as follows: $D_E$ = ETH-XGaze, $D_M$ = MPIIGaze, $D_G$ = Gaze360, and $D_D$ = EyeDiap.
Baseline is a ResNet-18 model without any fine-tuning.
Results are reported in angular gaze error ($^\circ$).
\textbf{Bold} indicates the best result.
The symbol $\dagger$ indicates results obtained from the re-implementation of TPGaze \cite{LiuQLHWPY24}.}
\label{tab:sota_cross_domain}
\end{table*}

\begin{table*}[t]
\centering
\setlength{\tabcolsep}{4pt}
\resizebox{0.82\textwidth}{!}{
\begin{tabular}{c|cc|ccc|cccc|c}
\toprule
\multirow{2}{*}{\textbf{Method}} & 
\multicolumn{2}{c|}{\textbf{Rank}} & 

\multicolumn{3}{c|}{\textbf{\# Parameters (M)}} & 
\multicolumn{4}{c|}{\textbf{Source $\rightarrow$ Target Domain (5-shot)}} & 
\multirow{2}{*}{\textbf{Avg}} \\
\cmidrule{2-10}
& \textbf{SVD} & \textbf{LoRA} & 
\textbf{Train} & \textbf{Tuned} & \textbf{Test} & 
\( D_E \mkern-4mu\rightarrow\mkern-4mu D_M \) & 
\( D_E \mkern-4mu\rightarrow\mkern-4mu D_D \) & 
\( D_G \mkern-4mu\rightarrow\mkern-4mu D_M \) & 
\( D_G \mkern-4mu\rightarrow\mkern-4mu D_D \) & \\

\midrule
Baseline (No Adaptation) & 64 & - & 2.31 & 0 & 2.31 & 6.60 & 8.84 & 6.86 & 6.83 & 7.29 \\
\hline
LoRA \cite{HuSWALWWC22} & 64 & 8 & 2.84 & 0.53 & 2.84 & 5.66 & 6.17 & 6.23 & 5.72 & 5.95 \\
MiLoRA \cite{WangLWCC25} & 64 & 8 & 2.84 & 0.29 & 2.84 & 5.67 & 6.25 & 6.23 & 6.00 & 6.04 \\
DoRA \cite{LiuWY0WCC24} & 64 & 8 & 2.87 & 0.56 & 2.87 & 5.51 & 5.83 & 6.30 & 5.98 & 5.91 \\
MoSLoRA \cite{WuW0W24} & 64 & 8 & 2.85 & 0.54 & 2.85 & 5.55 & 6.13 & 6.31 & \textbf{5.70} & 5.92 \\
MELoRA \cite{RenS0ZRRCP24} & 64 & 8 & 2.57 & 0.27 & 2.57 & 5.56 & 6.12 & 6.29 & 5.84 & 5.95 \\
Spectral Adapter$^A$ \cite{ZhangP24} & 64 & 8 & 2.59 & 0.29 & \textbf{2.31} & 5.50 & 6.15 & 6.23 & 6.00 & 5.97 \\
FLoRA \cite{SiW0XLDQ0025} & 64 & 8 & 2.38 & \textbf{0.07} & 2.38 & 5.85 & 6.36 & 6.40 & 5.71 & 6.08 \\
\hline
\rowcolor{orange!10}
\textbf{Alfa} & 64 & 8 & 5.26 & 2.98 & \textbf{2.31} & \textbf{5.30} & \textbf{5.82} & \textbf{5.91} & 5.86 & \textbf{5.72} \\
\bottomrule
\end{tabular}
}
\caption{
Comparison with other LoRA-based variant methods.
The baseline is \textit{not} fine-tuned on the target domain (\ie~no adaptation).
All methods use the same truncated pre-trained weight matrix $W_d = U_d S_d V_d^\top$.
While some variants introduce additional components that cannot be merged into $W_d$, Alfa supports full mergeability, enabling efficient inference without extra computational overhead.
\textbf{Bold} indicates the best result.
}
\label{tab:lora_variants_comp}
\end{table*}

\subsection{Datasets}
For gaze estimation, we use a cross-domain setup \cite{LiuQLHWPY24, BaoLWL22, LiuLWL21} with models trained on ETH-XGaze ($D_E$) or Gaze360 ($D_G$) as source domains, and evaluated on MPIIGaze ($D_M$) and EyeDiap ($D_D$) as target domains.
All preprocessing follows TPGaze \cite{LiuQLHWPY24}.
For LLM experiments, we fine-tune on the s1K reasoning dataset \cite{MYSLLHZLCH25} and evaluate zero-shot performance on GSM8K \cite{CKBCJKPTHNHS21}, MATH500 \cite{LightmanKBEBLLS24}, Countdown \cite{PZWYPS25}, and Sudoku \cite{Sudoku}.
Full dataset details are in the supplementary material.
\subsection{Experimental Setup}
We use one NVIDIA RTX 4090 for gaze experiments, and two NVIDIA A40 GPUs for LLM experiments.
For TTP pre-training and personalization, learning rate is set to $10^{-4}$.
For LLM adaptation, we follow the fine-tuning from d1 \cite{ZhaoGZG25} with LLaDA-8B-Instruct \cite{NieZYZOHZLWL25} as the base model.
Below, we describe the details for TTP:
\paragraph{Pre-training}
We use Adam with a batch size of 120 for 50 epochs on source domains $D_E$ and $D_G$.
After applying SVD, we fine-tune all parameters on the same data for 25 additional epochs to mitigate information loss, resulting in the truncated pre-trained weights $W_d$.
\paragraph{Personalization}
We use the first 5 images per subject as in TPGaze \cite{LiuQLHWPY24}.
Each image is repeated once and augmented with ColorJitter, GaussianBlur, and RandomAffine (see supplementary material for details).
We use AdamW with a 10$\times$ learning rate for layer3 and layer4 of ResNet-18, and apply the same fine-tuning scheme across all LoRA variants for fair comparison.
\paragraph{Evaluation Metric}
We report angular gaze error (in degrees), measuring the angle between predicted and ground-truth gaze directions.
We convert model outputs, 2D yaw and pitch, into 3D vectors to compute the error, following prior works \cite{LiuLWL21, BaoLWL22, LiuQLHWPY24}.

\begin{table*}[t]
\centering
\setlength{\tabcolsep}{4pt}
\resizebox{0.75\textwidth}{!}{
\begin{tabular}{c|c|c|C{2.5cm}|C{1.1cm}C{1.1cm}|C{1.3cm}C{1.3cm}|C{1.4cm}C{1.4cm}|C{1.0cm}C{1.0cm}}
\toprule
\multirow{2}{*}{\textbf{Backbone}} &
\multirow{2}{*}{\textbf{Method}} &
\multirow[b]{2}{*}{\shortstack[c]{\textbf{LoRA} \\ \textbf{Rank}}} &
\multirow[b]{2}{*}{\shortstack[c]{\textbf{Tuned Params} \\ (Usage \%)}} &
\multicolumn{2}{c|}{\textbf{GSM8K (0-shot)}} &
\multicolumn{2}{c|}{\textbf{MATH500 (0-shot)}} &
\multicolumn{2}{c|}{\textbf{Countdown (0-shot)}} &
\multicolumn{2}{c}{\textbf{Sudoku (0-shot)}} \\
\cmidrule(lr){5-6}
\cmidrule(lr){7-8}
\cmidrule(lr){9-10}
\cmidrule(lr){11-12}
& & & &
\textbf{128} & \textbf{256} &
\textbf{128} & \textbf{256} &
\textbf{128} & \textbf{256} &
\textbf{128} & \textbf{256} \\
\midrule
\multirow{4}{*}{LLaDA-8B-Instruct}
& LoRA & 128 & 100.7M (1.24 $\%$) & 66.5 & \textbf{78.8} & \textit{26.2} & 32.6 & 20.3 & 14.5 & \textit{16.5} & \textbf{8.5} \\
& DoRA$^\dagger$ & 128 & 101.1M (1.25$\%$) & \textit{68.1} & 76.8 & \textit{26.2} & \textit{33.4} & \textit{21.5} & \textit{16.0} & \textbf{17.2} & 8.0 \\
& LoRA$^\dagger$ & 64 & \textbf{50.3M (0.62$\%$)} & 67.9 & \textit{77.9} & 25.0 & 33.0 & 16.4 & 12.5 & 14.5 & 7.9 \\
& \cellcolor{orange!10}Alfa & \cellcolor{orange!10}64 & \cellcolor{orange!10}69.2M (0.85$\%$) &
\cellcolor{orange!10}\textbf{68.4} & \cellcolor{orange!10}77.1 &
\cellcolor{orange!10}\textbf{26.6} & \cellcolor{orange!10}\textbf{33.8} &
\cellcolor{orange!10}\textbf{27.3} & \cellcolor{orange!10}\textbf{17.2} &
\cellcolor{orange!10}9.7 & \cellcolor{orange!10}\textit{8.3} \\
\bottomrule
\end{tabular}
}
\caption{
Comparison of Alfa, LoRA, and DoRA on LLaDA-8B-Instruct across four zero-shot reasoning tasks.
Alfa achieves competitive or better performance while using only 0.85\% of tunable parameters.
\textbf{Bold} indicates the best result, and \textit{italics} indicate the second best.
The symbol $\dagger$ indicates re-implementation.
}
\label{tab:dllm}
\end{table*}

\subsection{Comparison with SOTA Methods}
Table \ref{tab:sota_cross_domain} compares Alfa with state-of-the-art TTP methods across four cross-domain gaze estimation benchmarks.
We evaluate model adaptation from ETH-XGaze or Gaze360 to MPIIGaze and EyeDiap.
The baseline is a ResNet-18 \cite{HeZRS16} model without fine-tuning.
For fair comparison, we also adopt ResNet-18 as the backbone.
Alfa achieves the lowest average angular gaze error across all benchmarks while remaining around $5\times$ smaller than other methods.

\subsection{Comparison with LoRA-based Methods}
We compare Alfa with other LoRA-based methods (see Section \ref{sec:related_work_lora}) in Table \ref{tab:lora_variants_comp}, and illustrate their architectural differences in Figure \ref{fig:lora_variants}.
All methods operate on the same truncated pre-trained weight $W_d = U_d S_d V_d^\top$.
However, not all LoRA-based variants support merging updates into this decomposed form of $W_d$ during inference.
In contrast, Alfa exploits the spatial structure of pre-trained weights and reweights semantically meaningful patterns.
Alfa's personalized updates are fully compatible with the truncated SVD form (see Section \ref{sec:trn_tst}), and incur no additional inference-time cost over other methods.
This gaze-specific structural prior guides adaptation toward meaningful filter reweighting without disrupting pre-trained semantics, leading to the lowest average gaze error across four cross-domain benchmarks.

\subsection{Ablation Studies}
We conduct ablation experiments to evaluate the effect of attention head count and the LoRA rank on Alfa's performance. 
For all settings, we fix SVD rank to 64.
As shown in Table \ref{tab:ab_alfa_heads}, increasing the number of attention heads generally improves personalization performance, dropping gaze error from 6.20 (1 head) to 5.72 (16 heads).
Since Alfa reuses the same left basis $U_d$ from the pre-trained $W_d$, all adaptations are merged into the base weight at inference time, incurring no additional computational cost regardless of the number of heads (see Section \ref{sec:trn_tst}).
Additional ablation studies are included in the supplementary materials.

\begin{figure}[!h]
\centering
\includegraphics[width=0.55\columnwidth]{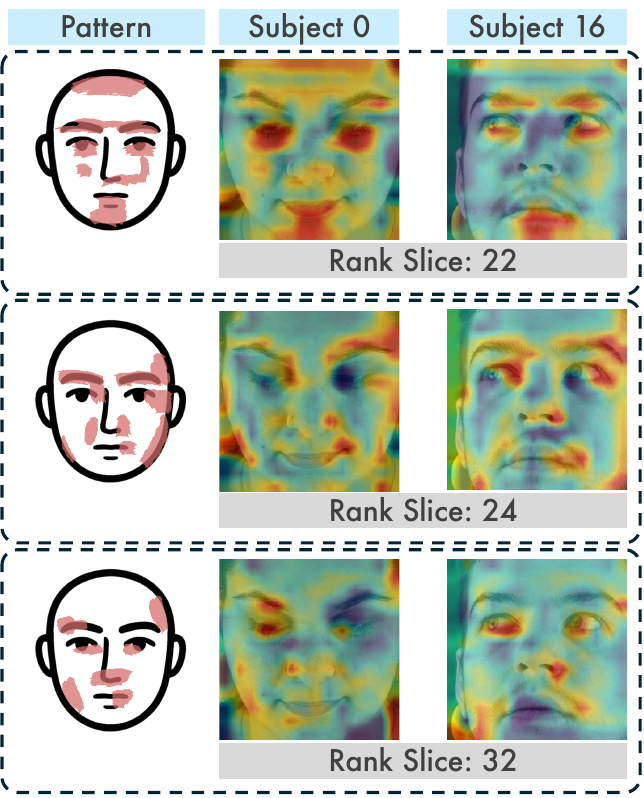}
\caption{
Spatial patterns captured during pre-training.
Visualizations use rank slices from SVD-decomposed weights of ResNet-18 (pre-trained on ETH-XGaze) from conv1 and the first block of layer3.
Left column: visualization of encoded pattern.
Middle and right columns: activations for Subject 0 and 16 from ETH-XGaze using $U_d[:, s] S_d[s] V_d^\top[s]$ for slice $s$.
Red regions indicate higher activations.
}
% \vspace{-9mm}
\label{fig:pattern_src_domain}
\end{figure}

\subsection{Visualization of Pre-trained Spatial Patterns}
\label{sec:vis_pretrain}
Figure \ref{fig:pattern_src_domain} illustrates some spatial patterns encoded during pre-training by visualizing individual rank slices from the SVD-decomposed weights of a ResNet-18 model trained on ETH-XGaze.
The $s$-th SVD slice is computed as:
\begin{equation}
    U_d[:, s] S_d[s] V_d^\top[s]
\end{equation}
where $s$ indexes the rank component of the decomposition.
The left column sketches the spatial structure encoded by that slice, while the middle and right columns display activations for two different subjects (Subject 0 and Subject 16).
For example, the 22nd slice emphasizes the eyebrows, lower eyelids, and lower mouth, while the 24th slice activates around the nose sides, regions beside the eyes, and facial muscles near the mouth.
Consistent patterns across individuals demonstrate that pre-training captures reusable spatial structures aligned with facial geometry relevant to gaze.

\subsection{Visualization of Adaptation Behavior}
\begin{figure}[t]
\centering
\includegraphics[width=0.55\columnwidth]{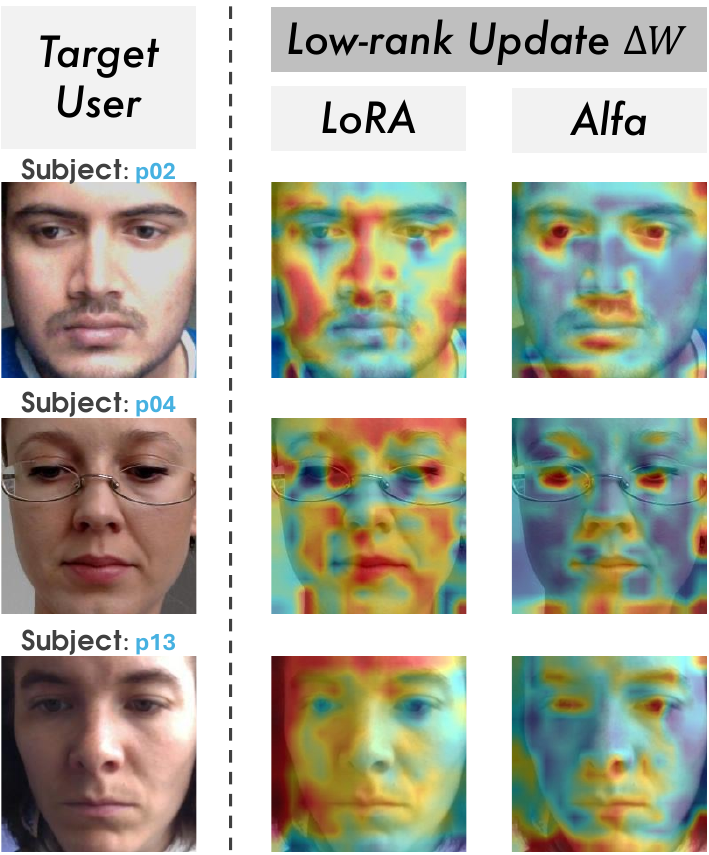}
\caption{
Visualization of low-rank updates $\Delta W$ on the MPIIGaze test set for LoRA and Alfa using filters from conv1 and the first block of layer3 in the ResNet-18 model (pre-trained on ETH-XGaze).
Red regions indicate higher activation values.
When using LoRA updates, model is highly inconsistent with respect to the significant regions of focus across users.
In contrast, Alfa captures localized regions consistently across users.
This shows Alfa identifies useful components that translate well between source and target domains from the semantic base dictionary. 
Reweighting these components allows for effective adaptation. 
}
\label{fig:lora_vs_alfa}
\end{figure}

We visualize low-rank updates $\Delta \mathcal{W}$ for three users from the MPIIGaze dataset (subjects p02, p04, and p13) in Figure \ref{fig:lora_vs_alfa}.
Alfa's updates consistently focus on gaze-relevant facial regions, such as the eyes and surrounding muscles.
The patterns vary slightly across users, reflecting personalized adjustments while still maintaining alignment with the model's original spatial semantics.
In contrast, LoRA yields dispersed and inconsistent updates even when using the same backbone layers with personalization.
However, we note that LoRA does not explicitly \textit{avoid} key regions: it simply lacks targeted semantic guidance and can rely on gaze-relevant signals (\eg, pose cues) from other regions.
This visualization highlights Alfa's ability to identify useful components from the semantic basis dictionary.
Since the components reflect domain-specific discrepancies between the pre-trained source model and the target user, we need only reweight them for effective user-specific adaptation.

\begin{table}[t]
\centering
\setlength{\tabcolsep}{3.5pt}
\resizebox{\columnwidth}{!}{

\begin{tabular}{c|ccc|cccc|c}
\toprule
\multirow{2}{*}{\textbf{Heads}} & 
% \multirow{2}{*}{\shortstack{\textbf{\#}\\\textbf{Heads}}} &  
\multicolumn{3}{c|}{\textbf{\# Parameters (M)}} & 
\multicolumn{4}{c|}{\textbf{Source $\mkern-2mu\rightarrow\mkern-2mu$ Target Domain (5-shot)}} &
\multirow{2}{*}{\textbf{Avg}} \\
\cmidrule{2-8}
& \textbf{Train} & \textbf{Tuned} & \textbf{Test} & 
\( D_E \mkern-4mu\rightarrow\mkern-4mu D_M \) & 
\( D_E \mkern-4mu\rightarrow\mkern-4mu D_D \) & 
\( D_G \mkern-4mu\rightarrow\mkern-4mu D_M \) & 
\( D_G \mkern-4mu\rightarrow\mkern-4mu D_D \) & \\
\midrule
1 & 2.35 & 0.04 & 2.31 & 5.74 & 6.84 & 6.36 & 5.87 & 6.20 \\
2 & 2.40 & 0.10 & 2.31 & 5.91 & 6.65 & 6.32 & \textbf{5.75} & 6.16\\
4 & 2.58 & 0.27 & 2.31 & 5.47 & 6.12 & 6.32 & 5.82 & 5.93 \\
8 & 3.16 & 0.86 & 2.31 & 5.67 & 6.19 & 6.24 & 5.76 & 5.97 \\
\rowcolor{orange!10}
16 & 5.26 & 2.98 & 2.31 & 5.30 & \textbf{5.82} & \textbf{5.91} & 5.86 & \textbf{5.72}\\
32 & 13.20 & 10.90 & 2.31 & \textbf{5.20} & 6.17 & 6.10 & 6.17 & 5.91\\
\bottomrule
\end{tabular}
}

\caption{
Ablation study on attention head count in Alfa.
Personalization performance generally improves with head count, with the lowest average gaze error at 16 heads. 
Heads share the left basis $U_d$, so adapted weights can be merged into the base model without extra computation at inference.
}
\label{tab:ab_alfa_heads}
\end{table}

\subsection{Applying Alfa to Diffusion-Based LLMs}
Table \ref{tab:dllm} compares Alfa to LoRA and DoRA when applied to the diffusion-based LLaDA-8B-Instruct model across four zero-shot reasoning benchmarks: GSM8K, MATH500, Countdown, and Sudoku.
We adapt with 1,000 samples.
We include LLM experiments as reasoning tasks reuse token-interaction patterns (\eg, formats, step markers), and reweighting these patterns is shown to be beneficial when data is limited.
LoRA results are from d1 \cite{ZhaoGZG25}.
We reimplement DoRA using HuggingFace PEFT library\footnote{\url{https://github.com/huggingface/peft}}.
While LoRA and DoRA use a LoRA rank of 128, Alfa uses a lower rank of 64 and tunes only 0.85\% of the model's parameters.
For all experiments in Table \ref{tab:dllm}, we retain the full pre-trained weight matrix $W$ without SVD truncation.
Alfa uses a SVD rank of 128 and 8 heads for computing the low-rank update $\Delta W$.
Despite this smaller footprint, Alfa achieves comparable or superior accuracy across benchmarks.
This suggests that reasoning patterns in language models may also be representable by generalizable components encoded during pre-training and of interest for future work.

\section{Conclusion}
We present Alfa, a structure-aware method for test-time personalization of gaze estimation models.
By attending over spatial patterns extracted via SVD, Alfa reuses meaningful components from pre-trained filters, enabling efficient domain adaptation through a multi-head low-rank design.
This approach allows scalable personalization during fine-tuning and maintains a compact model without increasing inference cost.
Experiments on four cross-domain gaze benchmarks demonstrate state-of-the-art performance with only a few unlabeled samples.
Furthermore, Alfa's structured adaptation shows promise for other applications, such as zero-shot reasoning tasks with diffusion-based language models.

\appendix

\section{TTP Datasets}
We use four gaze datasets in out experiments:
\begin{itemize}
    \item \textbf{ETH-XGaze} ($D_E$) \cite{ZhangPBBTH20} is a large-scale dataset with 756,540 images from 80 subjects, collected in a controlled indoor environment using a synchronized multi-view setup with varied illumination. It covers a broad range of head poses (up to about $\pm80^\circ$) and gaze directions (up to about $\pm120^\circ$), with dense sampling of gaze targets. These characteristics provide consistent annotations across substantial appearance and pose variation. We use ETH-XGaze as a source domain for pre-training.
    \item \textbf{Gaze360} ($D_G$) \cite{KellnhoferRSM019} contains 100,933 images after filtering out samples without visible faces, following prior work \cite{LiuLWL21, LiuQLHWPY24}. This dataset is recorded in unconstrained indoor and outdoor environments, with large variability in lighting, backgrounds, and viewing distances. It covers a very wide range of head poses (up to about $\pm90^\circ$) and gaze directions (up to about $\pm140^\circ$), reflecting physically unconstrained viewing conditions. These characteristics provide strong appearance diversity that complements controlled datasets. We use Gaze360 as a source domain for pre-training.
    \item \textbf{MPIIGaze} ($D_M$) \cite{ZhangSFB19} contains gaze data collected from subjects using personal laptops in everyday environments. This dataset reflects natural variation in appearance, illumination, and background, and includes moderate head-pose changes (up to about $\pm15^\circ$) and gaze directions (up to about $\pm20^\circ$) that commonly occur during daily laptop use. We follow the standard evaluation split \cite{ZhangSFB19}, which uses 3,000 images per subject. For adaptation, we use the first 5 images from each subject, following \cite{LiuQLHWPY24}.
    \item \textbf{EyeDiap} ($D_D$) \cite{MoraMO14} provides gaze recordings captured in controlled indoor settings across several session types, including variations in visual targets and head-pose conditions. This dataset covers moderate head-pose variation (up to about $\pm15^\circ$) and gaze directions (up to about $\pm25^\circ$). We use 16,674 images from 14 subjects under screen-target sessions as the evaluation set, and use the first 5 images per subject for adaptation, following \cite{LiuQLHWPY24}.
\end{itemize}

\section{LLM Reasoning Datasets and Evaluation}
The \textbf{s1K} dataset contains 1,000 curated examples of multi-format reasoning questions.
We first fine-tune LLaDA-8B Instruct on the s1K dataset for 20 epochs using a sequence length of 4096 tokens.
Then, we evaluate adaptation performance on four zero-shot reasoning tasks across mathematical and planning domains:
\begin{itemize}
    \item \textbf{GSM8K} consists of grade school math word problems that require multi-step arithmetic reasoning.
    \item \textbf{MATH500} is a curated subset of 500 high school competition math problems from the MATH dataset \cite{HendrycksBKABTS21}.
    \item \textbf{Sudoku} involves solving 4×4 puzzles through constraint-based number placement.
    \item \textbf{Countdown} is a numerical planning task where models must reach a target value using basic arithmetic operations on three given numbers.
\end{itemize}

\section{Data Augmentation for TTP}
To enhance model robustness, we apply data augmentation to the target user samples. 
Since only five images per subject are available, each image is repeated once and augmented to simulate realistic appearance variations.
We use PyTorch's torchvision library to apply three types of augmentations:
\begin{enumerate}
\item \textbf{ColorJitter} (adjusting brightness, contrast, saturation, and hue) to simulate lighting variations.
\item \textbf{GaussianBlur} to mimic camera shifts or softening effects caused by misfocus, motion, or lens variability. We use a larger kernel size (\ie, 5) for ETH-XGaze due to its controlled lab setting, to reflect real-world imperfections.
\item \textbf{RandomAffine} to introduce minor spatial distortions.
\end{enumerate}
Table \ref{tab:supp_aug} summarizes the augmentation settings used for each source–target adaptation pair. 
These augmentation settings are applied across all LoRA variants for fair comparison.

\begin{table*}[t]
\centering
\setlength{\tabcolsep}{4pt}
\resizebox{0.55\textwidth}{!}{
\begin{tabular}{l|cccc}
\toprule
\multirow{2}{*}{Augmentation} & \multicolumn{2}{c}{From $D_E$} & \multicolumn{2}{c}{From $D_G$} \\
\cmidrule(lr){2-3} \cmidrule(lr){4-5}
 & $\rightarrow D_M$ & $\rightarrow D_E$ & $\rightarrow D_M$ & $\rightarrow D_E$ \\
\midrule
Repeat & 1 & 1 & 1 & 1 \\
\hline
\multirow{4}{*}{ColorJitter} 
& brightness=0.2 & brightness=0.2 & brightness=0.2 & brightness=0.2 \\
& contrast=0.2   & contrast=0.2   & contrast=0.2   & contrast=0.2 \\
& saturation=0.1 & saturation=0.1 & saturation=0.1 & saturation=0.1 \\
& hue=0.05       & hue=0.05       & hue=0.05       & hue=0.05 \\
\hline
GaussianBlur & kernel\_size=5 & kernel\_size=5 & kernel\_size=3 & kernel\_size=3 \\
\hline
\multirow{2}{*}{RandomAffine} & degrees=5 & degrees=5 & degrees=5 & degrees=5 \\
& translate=(0.02, 0.02) & translate=(0.02, 0.02) & translate=(0.02, 0.02) & translate=(0.02, 0.02) \\
\bottomrule
\end{tabular}
}
\caption{
Gaze data augmentation settings used for TTP.
We use PyTorch \texttt{torchvision} to apply augmentations to each target user image.
Each image is repeated once to generate 10 personalized samples per subject.}
\label{tab:supp_aug}
\end{table*}

\section{Additional Results on SOTA Comparisons}
We present additional comparisons with source-available UDA methods (\ie, where source-domain data is available) in Table \ref{tab:supp_sota}, including DAGEN \cite{GuoYZCLZ20}, GazeAdv \cite{WangZSJ19}, and Gaze360 \cite{KellnhoferRSM019}.
Alfa outperforms these methods despite lacking access to source-domain data.
We also include the result from a source-available in-domain supervised method,  DFT Gaze \cite{HsiehLZTCSLK24}, which adapts using labeled data from the target domain.
Even with supervision from DFT Gaze, Alfa achieves lower gaze error when adapting to $D_M$, demonstrating the effectiveness of its structure-aware personalization approach.

\begin{table*}[t!]
\centering
\setlength{\tabcolsep}{4pt}
\resizebox{0.71\textwidth}{!}{
\begin{tabular}{c|ccc|cccc|c}
\toprule
\multirow{2}{*}{\textbf{Method}} & 
\multicolumn{3}{c|}{\textbf{\# Parameters (M)}} & 
\multicolumn{4}{c|}{\textbf{Source $\rightarrow$ Target Domain (5-shot)}} & 
\multirow{2}{*}{\textbf{Avg}} \\
\cmidrule{2-8}
&
\textbf{Train} & \textbf{Tuned} & \textbf{Test} & 
\( D_E \rightarrow D_M \) & 
\( D_E \rightarrow D_D \) & 
\( D_G \rightarrow D_M \) & 
\( D_G \rightarrow D_D \) & \\
\midrule
Baseline (ResNet-18) & 11.18 & 0 & 11.18 & 8.02 & 7.30 & 7.79 & 8.19 & 7.83 \\
\hline
\rowcolor{gray!10}
\multicolumn{9}{l}{\textbf{Source-available UDA Methods}} \\
DAGEN \cite{GuoYZCLZ20} & - & - & - & 7.53 & 8.46 & 9.31 & 12.05 & 9.34 \\
GazeAdv \cite{WangZSJ19} & - & - & - & 8.48 & 7.70 & 9.15 & 11.15 & 9.12 \\
Gaze360 \cite{KellnhoferRSM019} & - & - & - & 7.86 & 9.64 & 7.71 & 9.54 & 8.69 \\
\hline
\rowcolor{gray!10}
\multicolumn{9}{l}{\textbf{TTP Methods}} \\
PnP-GA$^\dagger$ \cite{LiuLWL21} & 116.9 & 116.9 & 116.9 & 6.91 & 7.18 & 7.36 & 8.17 & 7.41 \\
RUDA$^\dagger$ \cite{BaoLWL22} & 12.20 & 12.20 & 12.20 & 6.86 & 6.84 & 6.96 & 5.32 & 6.50 \\
TPGaze \cite{LiuQLHWPY24} & 11.18 & 0.13 & 11.18 & 6.30 & 5.89 & 6.62 & \textbf{5.04} & 5.96 \\
\rowcolor{orange!10}
\textbf{Alfa} & 5.26 & 2.98 & 2.31 & \textbf{5.30} & \textbf{5.82} & \textbf{5.91} & 5.86 & \textbf{5.72} \\
\hline
\hline
\rowcolor{gray!10}
\multicolumn{9}{l}{\textbf{Source-available In-Domain Adaptation with Supervision}} \\
Method &
\textbf{Train} & \textbf{Tuned} & \textbf{Test} & 
\( D_M \rightarrow D_M \) & 
& & & \\
\hline
DFT Gaze \cite{HsiehLZTCSLK24} & 0.28 & \textbf{0.014} & \textbf{0.28} & 5.35 & - & - & - & - \\
\bottomrule
\end{tabular}
}
\caption{
Comparison with state-of-the-art source-available UDA and TTP methods across four cross-domain benchmarks. 
Results are reported in angular gaze error ($^\circ$).
Datasets are denoted: $D_E$ = ETH-XGaze, $D_M$ = MPIIGaze, $D_G$ = Gaze360, and $D_D$ = EyeDiap.
Baseline is a ResNet-18 model without any fine-tuning.
\textbf{Bold} indicates the best result.
The symbol $\dagger$ indicates results obtained from a re-implementation of TPGaze \cite{LiuQLHWPY24}.
We also include source-available in-domain adaptation results with supervision (\eg, DFT Gaze \cite{HsiehLZTCSLK24}).
Despite DFT Gaze having access to target-domain labels, Alfa achieves better performance when adapting to $D_M$ without labels.
}
\label{tab:supp_sota}
\end{table*}

\section{Visualization of Adaptation Behavior}
\begin{figure*}[!h]
\centering
\includegraphics[width=0.95\textwidth]{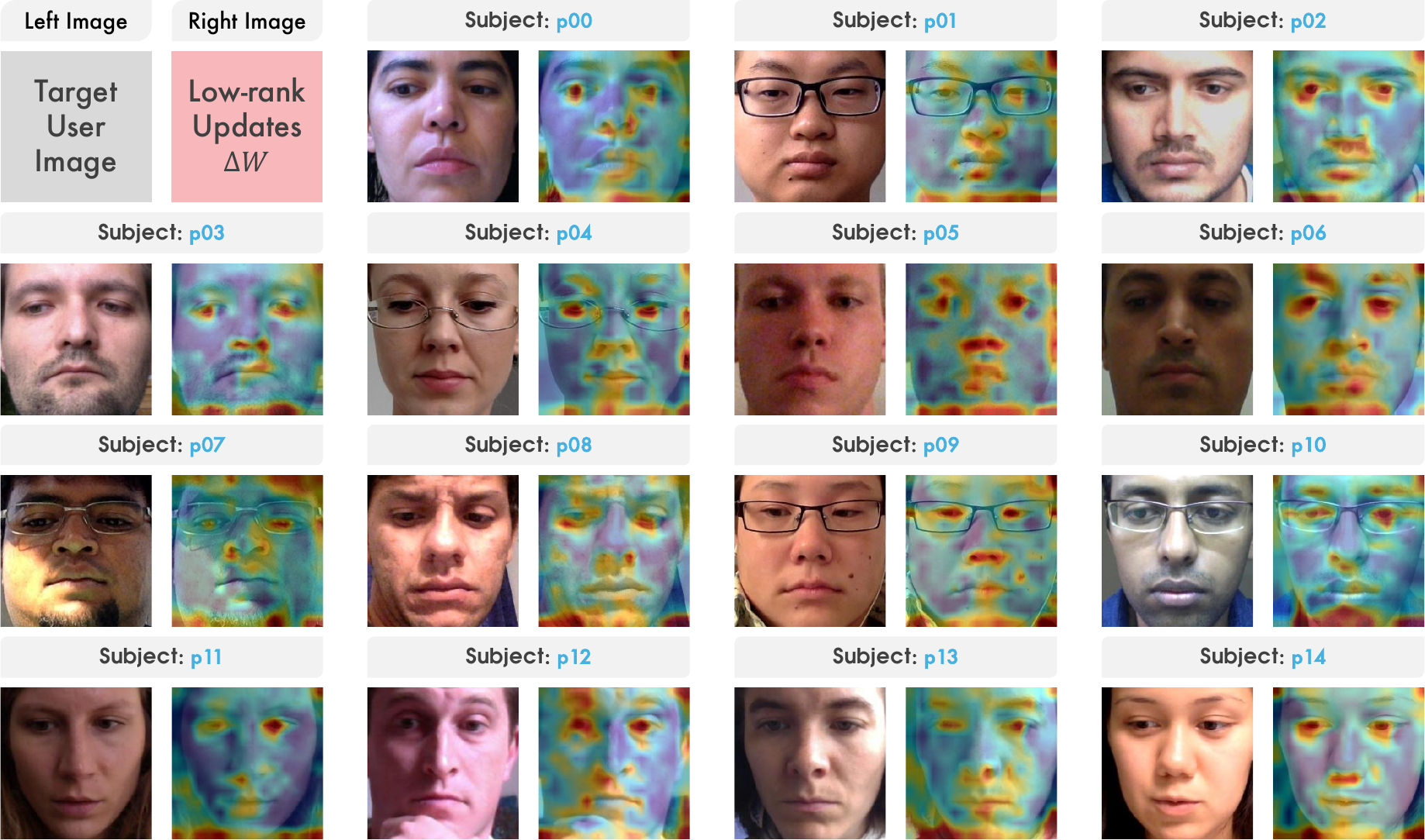}
\caption{
Visualization of low-rank updates $\Delta W$ across all target users on the MPIIGaze test set for \textit{Alfa}, using filters from conv1 and the first block of layer3 in a ResNet-18 model pre-trained on ETH-XGaze.
Red regions indicate stronger activation.
}
\label{fig:supp_vis_alfa_deltaW}
\end{figure*}

\begin{figure*}[!h]
\centering
\includegraphics[width=0.95\textwidth]{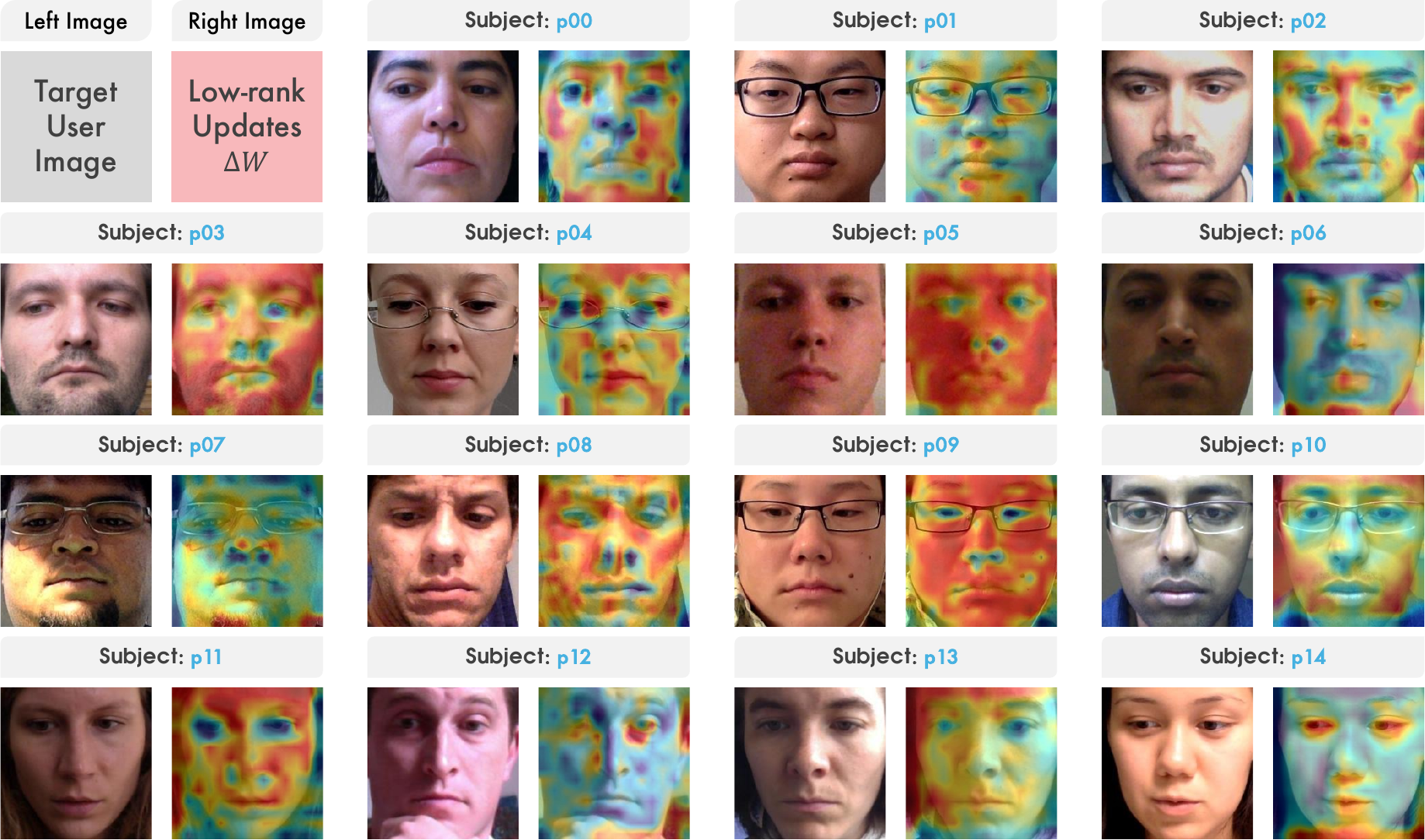}
\caption{
Visualization of low-rank updates $\Delta W$ across all target users on the MPIIGaze test set for \textit{LoRA}, using filters from conv1 and the first block of layer3 in a ResNet-18 model pre-trained on ETH-XGaze.
Red regions indicate stronger activation.
}
\label{fig:supp_vis_lora_deltaW}
\end{figure*}

In Figure 5 of the main paper, we visualize low-rank updates $\Delta W$ from Alfa and LoRA for three target users (p02, p04, and p13) on the MPIIGaze dataset.
We extend this visualization to all target users for Alfa in Figure \ref{fig:supp_vis_alfa_deltaW}, while Figure \ref{fig:supp_vis_lora_deltaW} shows the corresponding updates from LoRA.

We compare low-rank updates from Alfa and LoRA across all target users in Figures \ref{fig:supp_vis_alfa_deltaW} and \ref{fig:supp_vis_lora_deltaW}.
LoRA produces dispersed and less structured changes, while Alfa yields more spatially coherent and gaze-aligned updates.
This highlights the effectiveness of Alfa's attention-based reweighting over the semantic basis dictionary.

\section{Visualization of Attention over Rank Slices}
\label{sec:supp_vis_rank_slice}
\begin{figure*}[!h]
\centering
\includegraphics[width=0.95\textwidth]{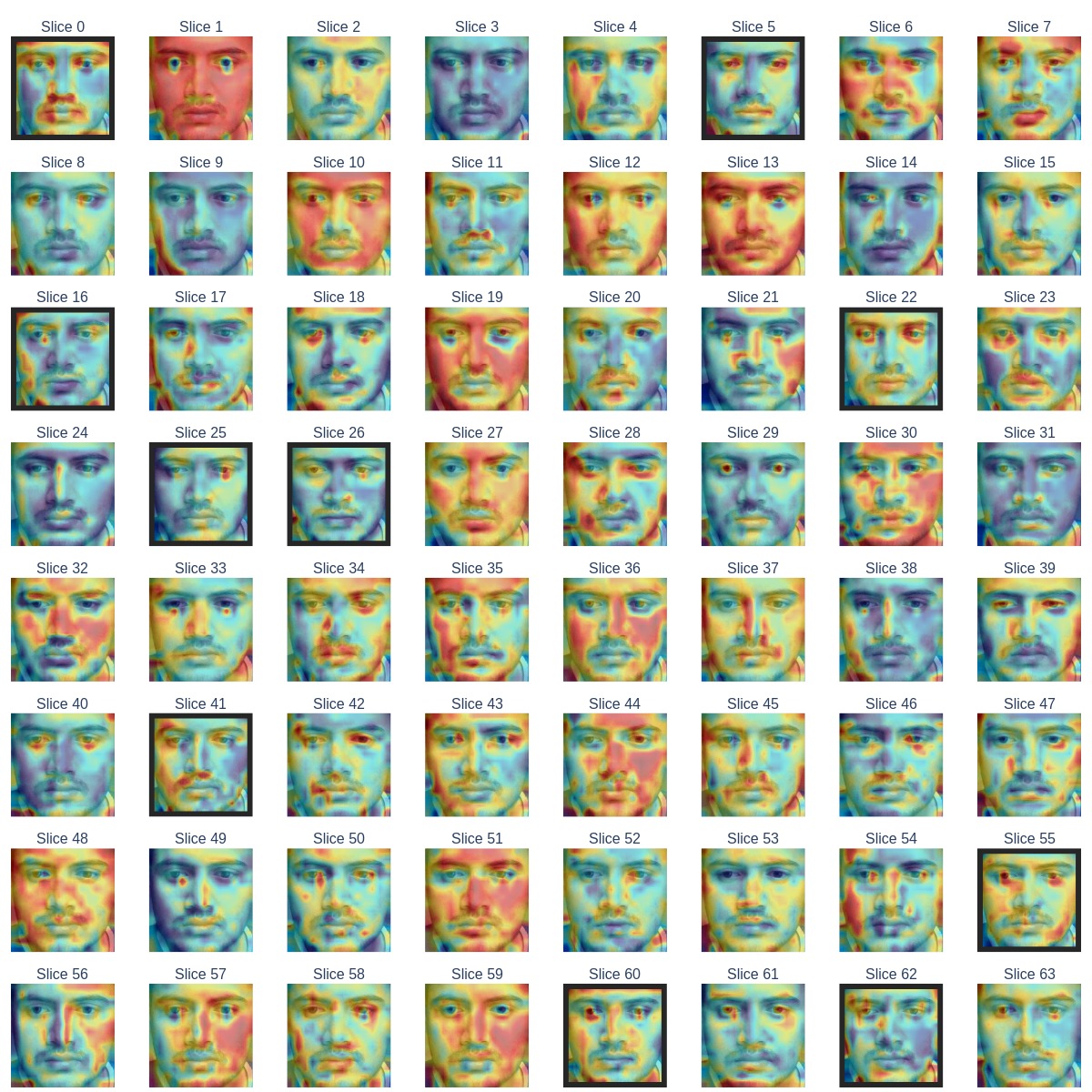}
\caption{
Visualization of selected rank slices from Alfa with head index $h = 0$ for subject \textit{p02} on the MPIIGaze test set.
We show 64 rank slices from conv1 and the first block of layer3 in a ResNet-18 model pre-trained on ETH-XGaze.
Red regions indicate stronger activation.
Black borders highlight the top-10 personalized rank slices selected for subject \textit{p02}.
Head index starting from 0.
}
\label{fig:supp_vis_p02_head0}
\end{figure*}

\begin{figure*}[!h]
\centering
\includegraphics[width=0.95\textwidth]{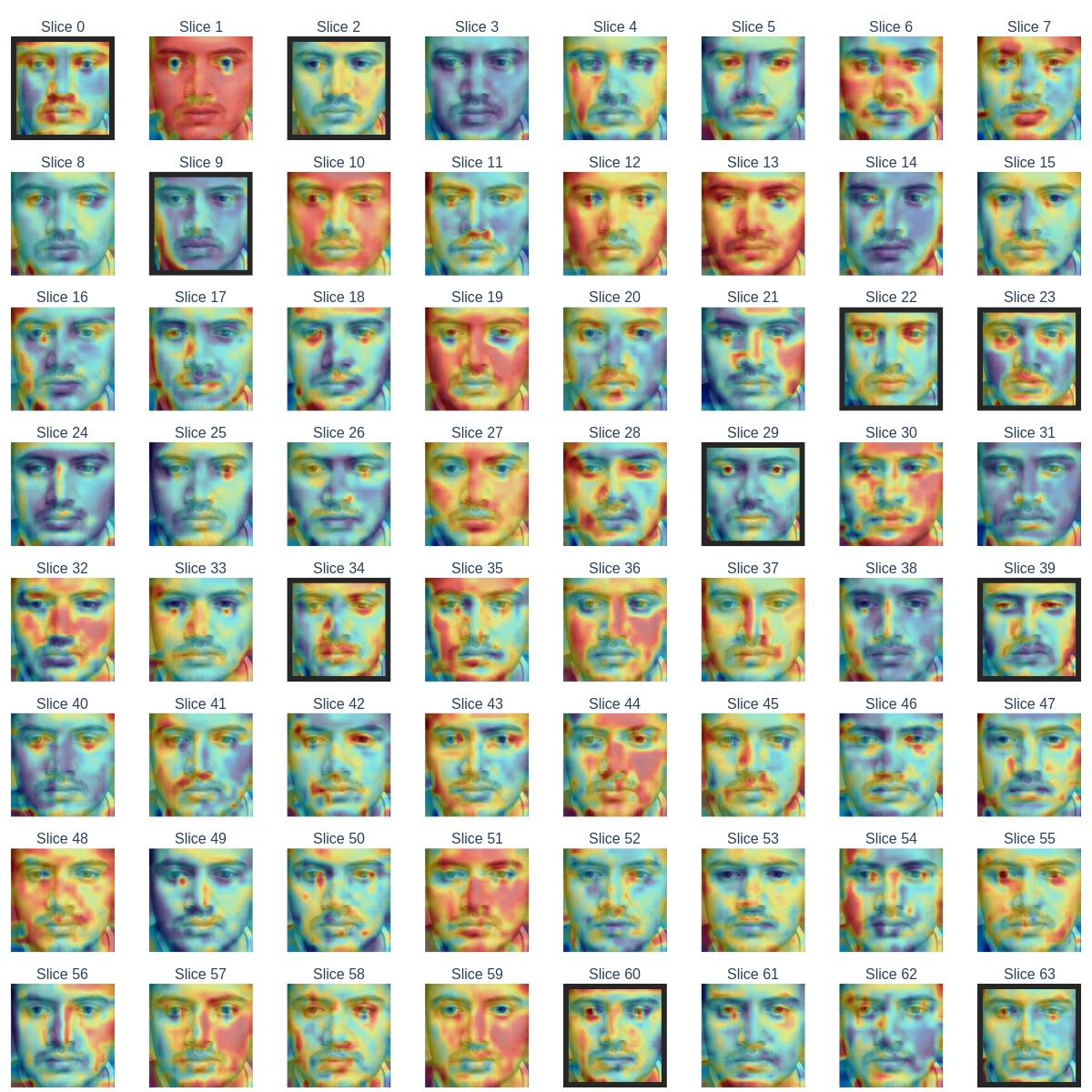}
\caption{
Visualization of selected rank slices from Alfa with head index $h = 7$ for subject \textit{p02} on the MPIIGaze test set.
We show 64 rank slices from conv1 and the first block of layer3 in a ResNet-18 model pre-trained on ETH-XGaze.
Red regions indicate stronger activation.
Black borders highlight the top-10 personalized rank slices selected for subject \textit{p02}.
Head index starting from 0.
}
\label{fig:supp_vis_p02_head7}
\end{figure*}

\begin{figure*}[!h]
\centering
\includegraphics[width=0.95\textwidth]{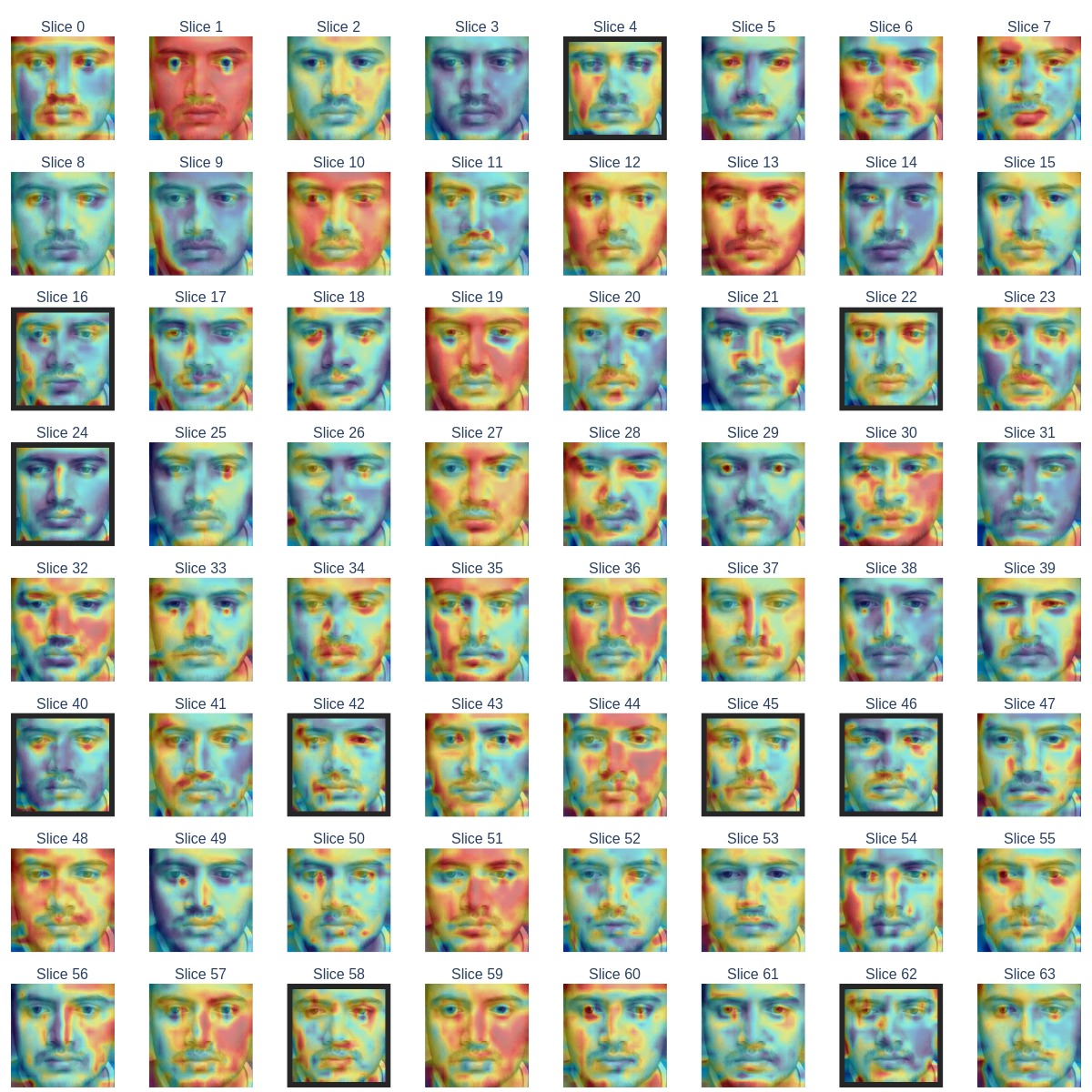}
\caption{
Visualization of selected rank slices from Alfa with head index $h = 12$ for subject \textit{p02} on the MPIIGaze test set.
We show 64 rank slices from conv1 and the first block of layer3 in a ResNet-18 model pre-trained on ETH-XGaze.
Red regions indicate stronger activation.
Black borders highlight the top-10 personalized rank slices selected for subject \textit{p02}.
Head index starting from 0.
}
\label{fig:supp_vis_p02_head12}
\end{figure*}

\begin{figure*}[!h]
\centering
\includegraphics[width=0.95\textwidth]{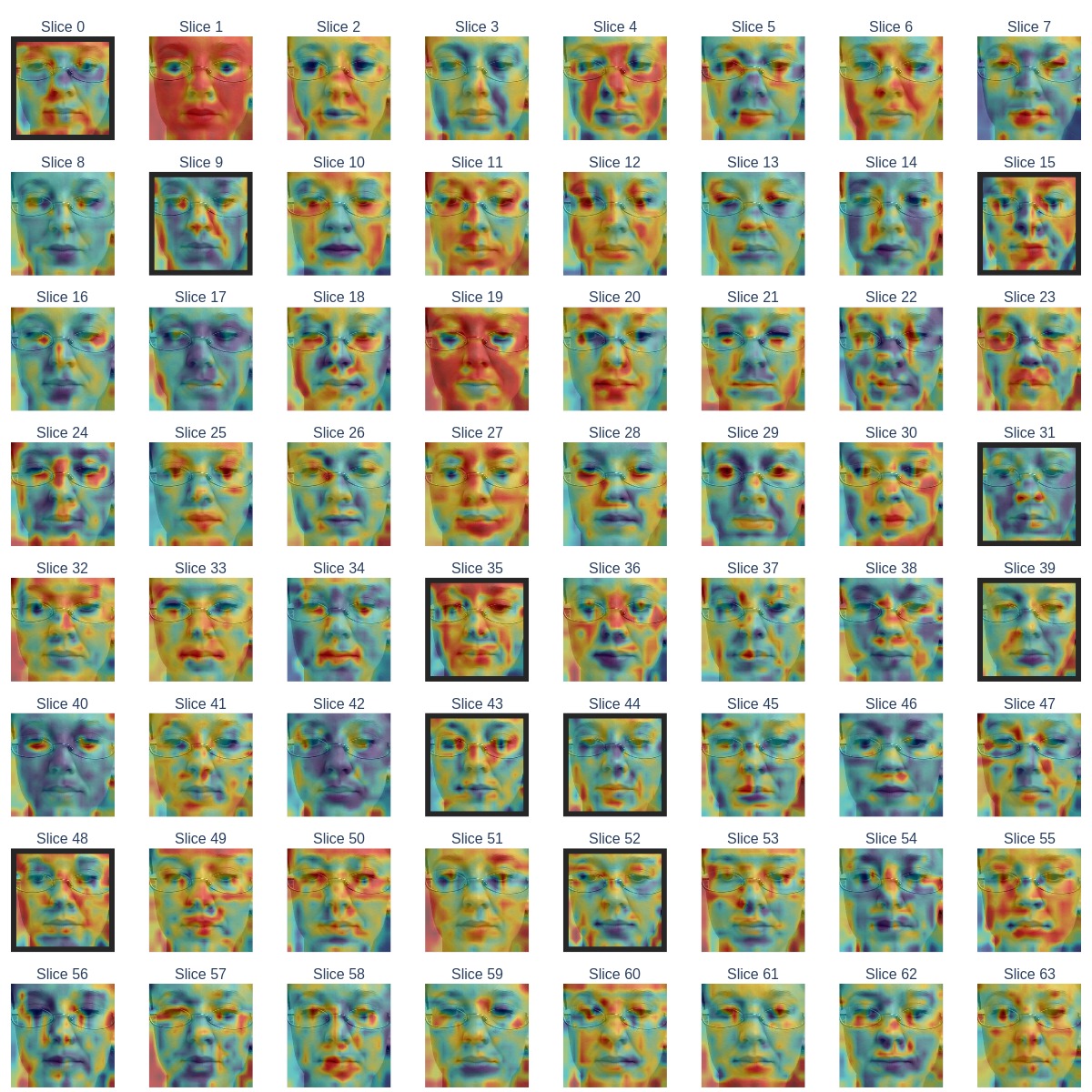}
\caption{
Visualization of selected rank slices from Alfa with head index $h = 1$ for subject \textit{p04} on the MPIIGaze test set.
We show 64 rank slices from conv1 and the first block of layer3 in a ResNet-18 model pre-trained on ETH-XGaze.
Red regions indicate stronger activation.
Black borders highlight the top-10 personalized rank slices selected for subject \textit{p04}.
Head index starting from 0.
}
\label{fig:supp_vis_p04_head1}
\end{figure*}

\begin{figure*}[!h]
\centering
\includegraphics[width=0.95\textwidth]{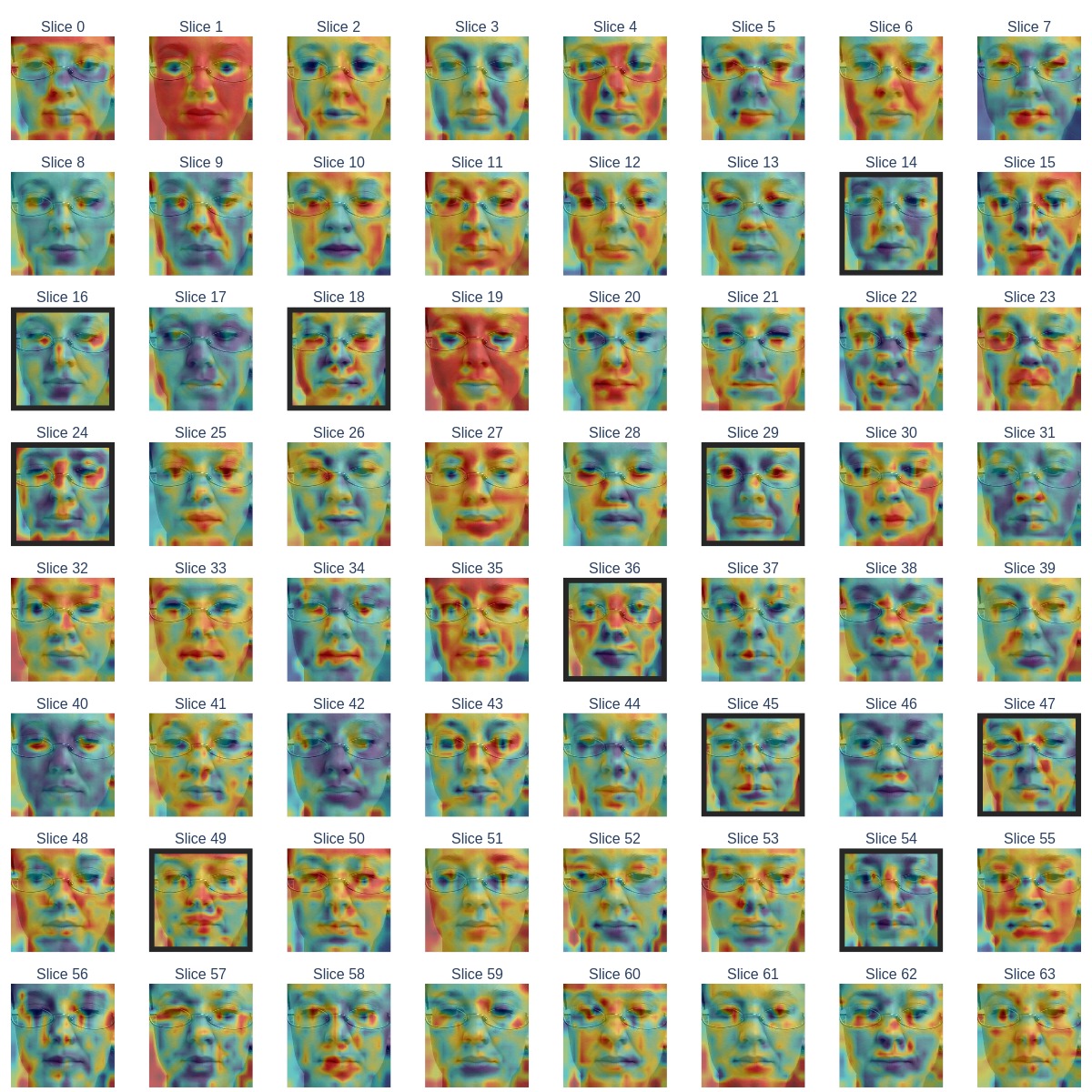}
\caption{
Visualization of selected rank slices from Alfa with head index $h = 4$ for subject \textit{p04} on the MPIIGaze test set.
We show 64 rank slices from conv1 and the first block of layer3 in a ResNet-18 model pre-trained on ETH-XGaze.
Red regions indicate stronger activation.
Black borders highlight the top-10 personalized rank slices selected for subject \textit{p04}.
Head index starting from 0.
}
\label{fig:supp_vis_p04_head4}
\end{figure*}

\begin{figure*}[!h]
\centering
\includegraphics[width=0.95\textwidth]{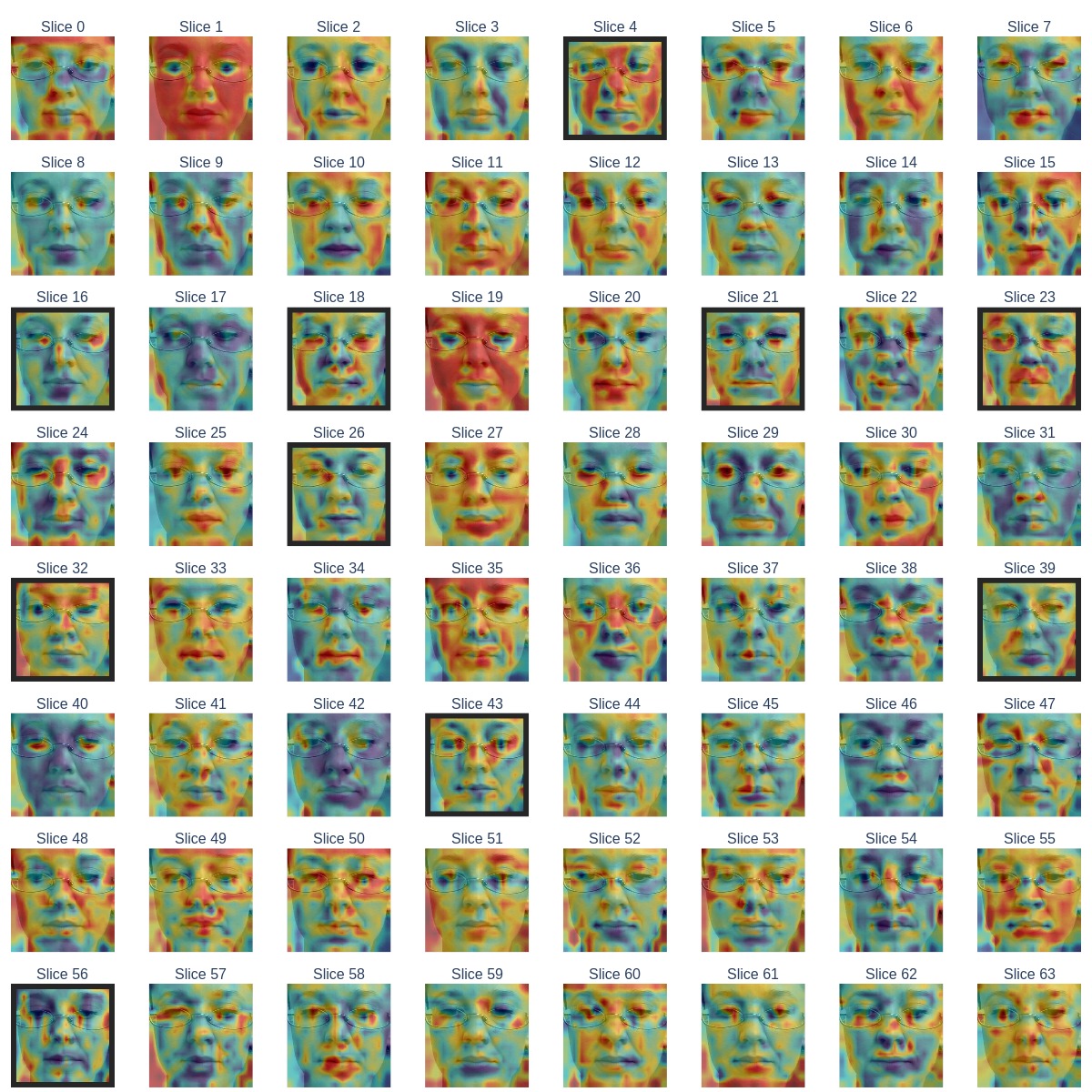}
\caption{
Visualization of selected rank slices from Alfa with head index $h = 10$ for subject \textit{p04} on the MPIIGaze test set.
We show 64 rank slices from conv1 and the first block of layer3 in a ResNet-18 model pre-trained on ETH-XGaze.
Red regions indicate stronger activation.
Black borders highlight the top-10 personalized rank slices selected for subject \textit{p04}.
Head index starting from 0.
}
\label{fig:supp_vis_p04_head10}
\end{figure*}

\begin{figure*}[!h]
\centering
\includegraphics[width=0.95\textwidth]{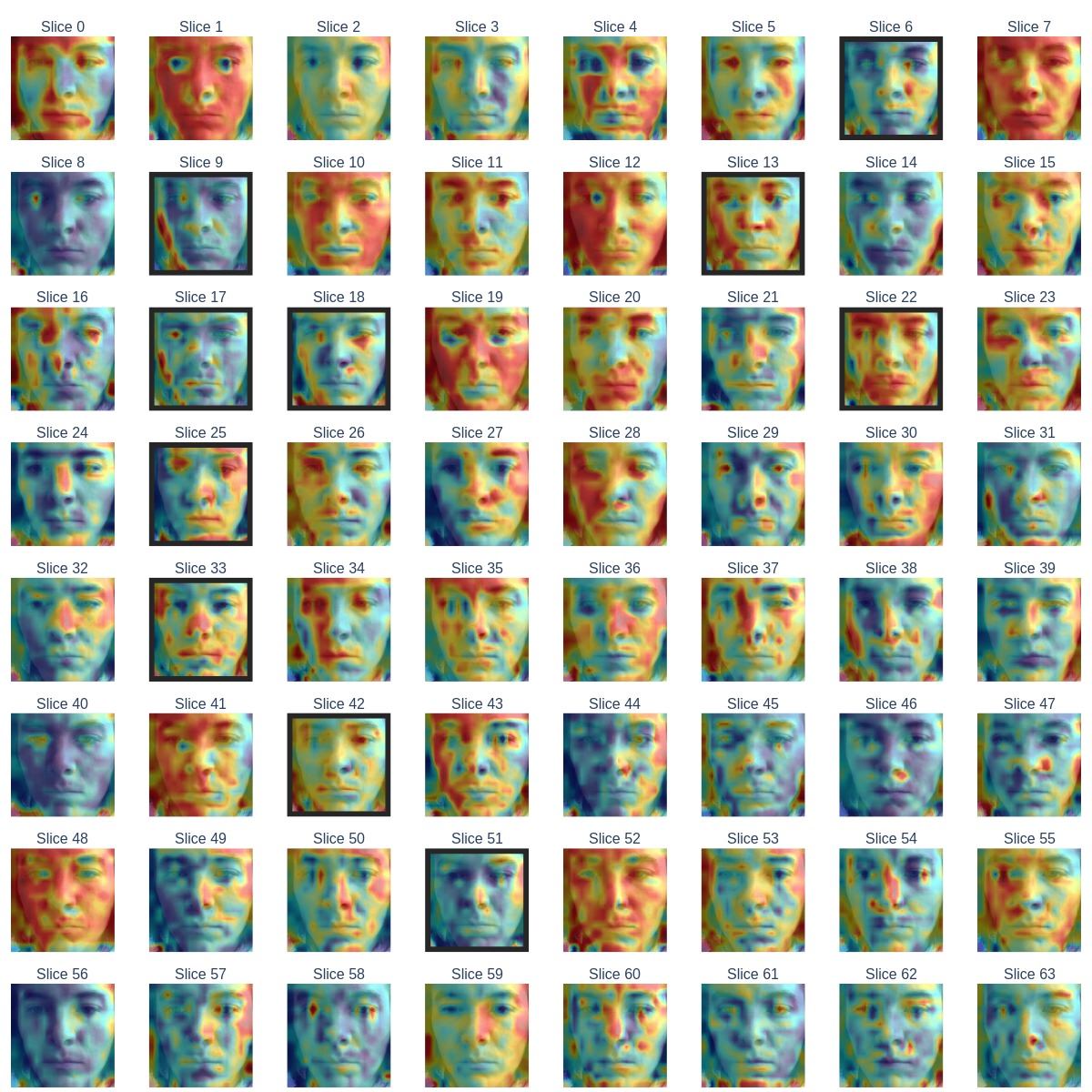}
\caption{
Visualization of selected rank slices from Alfa with head index $h = 5$ for subject \textit{p13} on the MPIIGaze test set.
We show 64 rank slices from conv1 and the first block of layer3 in a ResNet-18 model pre-trained on ETH-XGaze.
Red regions indicate stronger activation.
Black borders highlight the top-10 personalized rank slices selected for subject \textit{p13}.
Head index starting from 0.
}
\label{fig:supp_vis_p13_head5}
\end{figure*}

\begin{figure*}[!h]
\centering
\includegraphics[width=0.95\textwidth]{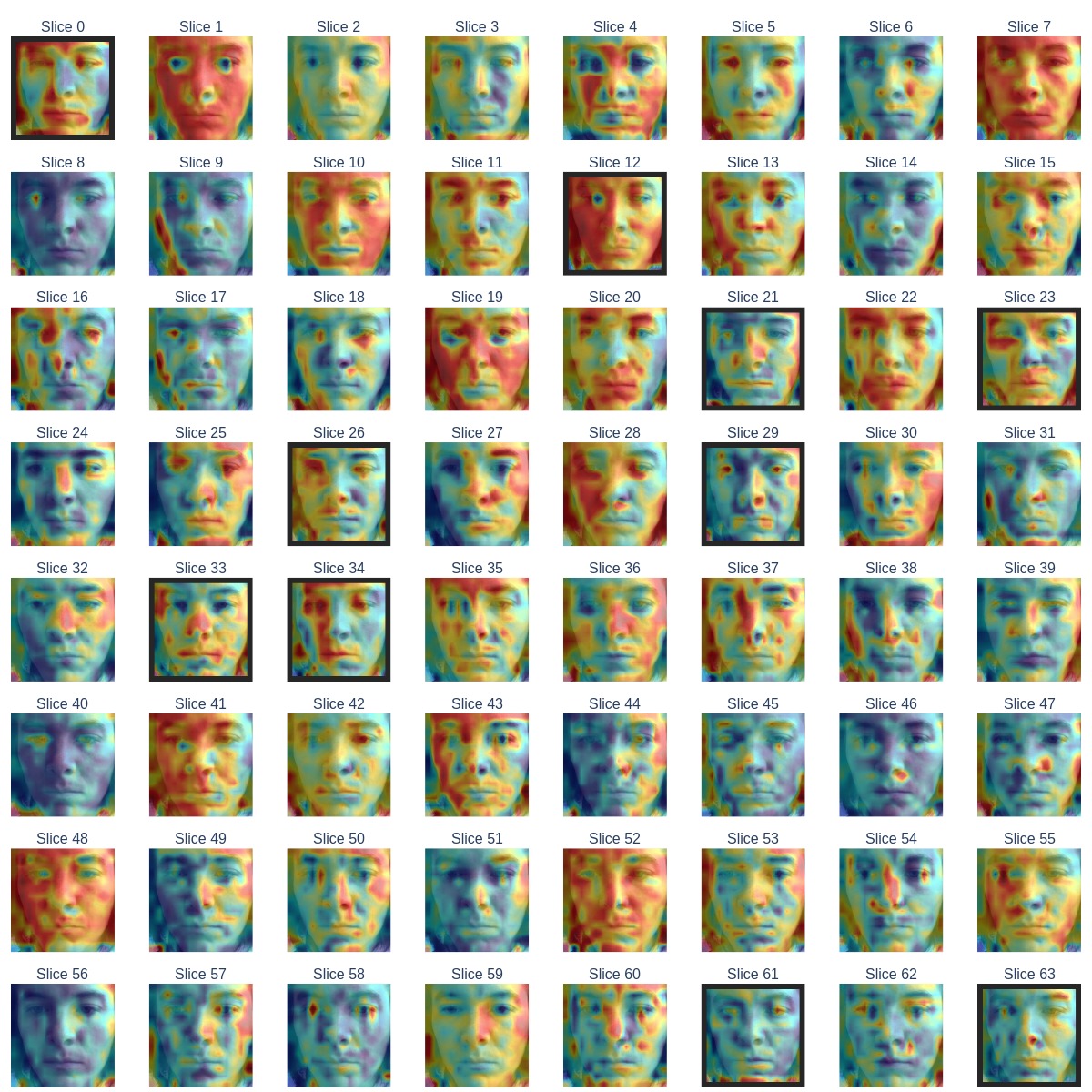}
\caption{
Visualization of selected rank slices from Alfa with head index $h = 6$ for subject \textit{p13} on the MPIIGaze test set.
We show 64 rank slices from conv1 and the first block of layer3 in a ResNet-18 model pre-trained on ETH-XGaze.
Red regions indicate stronger activation.
Black borders highlight the top-10 personalized rank slices selected for subject \textit{p13}.
Head index starting from 0.
}
\label{fig:supp_vis_p13_head6}
\end{figure*}

\begin{figure*}[!h]
\centering
\includegraphics[width=0.95\textwidth]{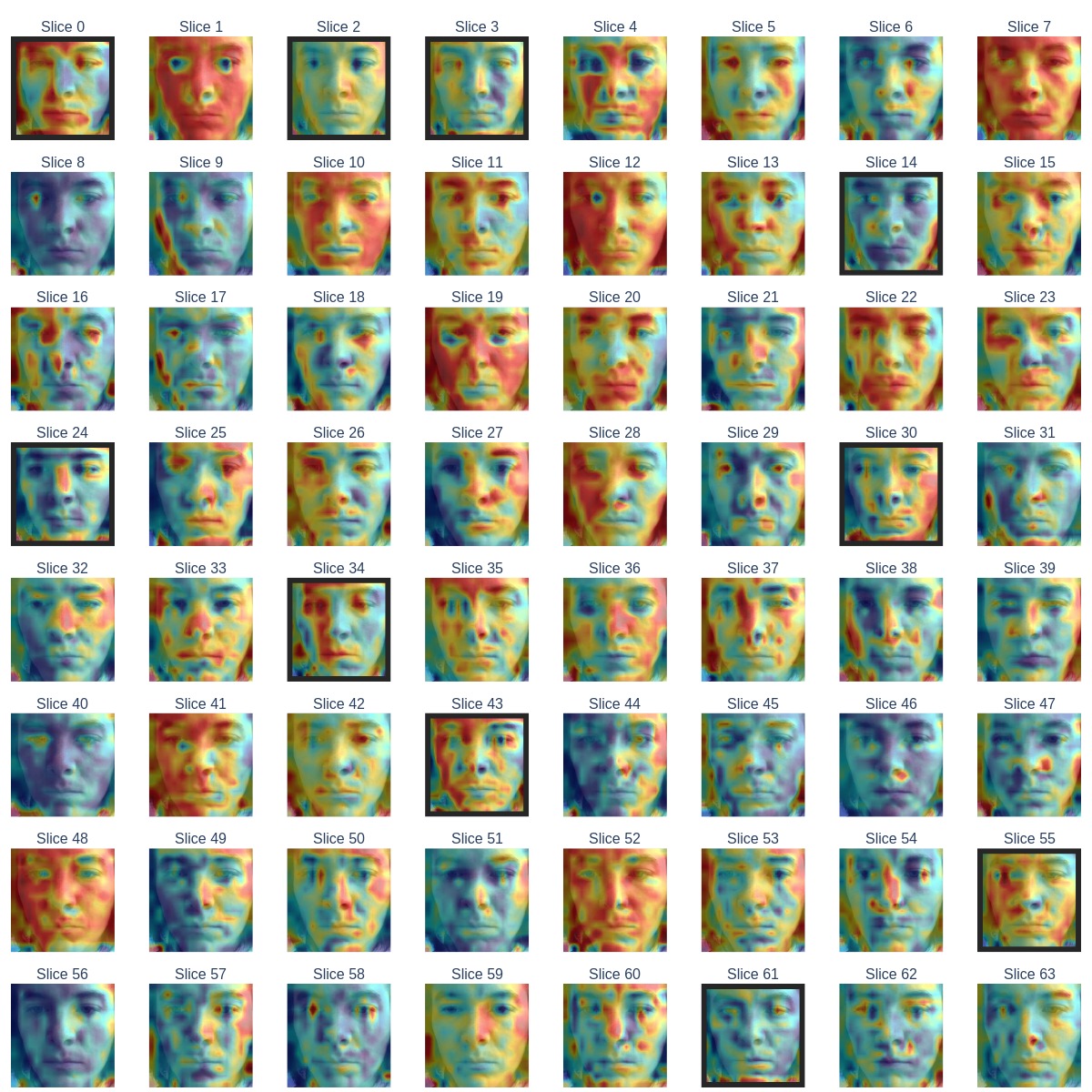}
\caption{
Visualization of selected rank slices from Alfa with head index $h = 9$ for subject \textit{p13} on the MPIIGaze test set.
We show 64 rank slices from conv1 and the first block of layer3 in a ResNet-18 model pre-trained on ETH-XGaze.
Red regions indicate stronger activation.
Black borders highlight the top-10 personalized rank slices selected for subject \textit{p13}.
Head index starting from 0.
}
\label{fig:supp_vis_p13_head9}
\end{figure*}

We visualize the behavior of individual attention heads over the semantic basis dictionary.
For each subject, we show how different heads selects different spatial patterns. Subject p02: Figures \ref{fig:supp_vis_p02_head0} to \ref{fig:supp_vis_p02_head12} show head 0, head 7, and head 12.
Subject p04: Figures \ref{fig:supp_vis_p04_head1} to \ref{fig:supp_vis_p04_head10} show head 1, head 4, and head 10.
Subject p13: Figures \ref{fig:supp_vis_p13_head5} to \ref{fig:supp_vis_p13_head9} show head 5, 6, and head 9.

The visualizations provide insight into how Alfa performs personalized adaptation.
We see that Alfa applies structured updates: each attention head selectively emphasizes different spatial patterns, aligning with gaze-relevant regions such as eye corners, eyelids, and facial muscles around the eyes.
The attention is distributed unevenly, which suggests the heads specialize in capturing different aspects of user-specific gaze features.
The final low-rank updates from Alfa for these three subjects are shown in Figure 5 of the main paper.

\section{Further Ablation Studies}
\label{sec:supp_add_ablation}

We provide additional ablation studies to examine how different design choices influence Alfa’s performance.
Table \ref{tab:ab_alfa_lora_ranks} shows a LoRA rank of 8 achieves the lowest average gaze error.
Table~\ref{tab:ab_alfa_svd_ranks} evaluates different SVD ranks and shows that a moderate rank of 64 achieves the best balance between adaptation capacity and stability.
Table~\ref{tab:ab_alfa_num_samples} analyzes the effect of the number of personal samples and indicates that performance improves with more images but saturates around 5-10 samples.
Table~\ref{tab:ab_alfa_noAttn} studies the role of the attention mechanism and demonstrates that removing attention degrades performance, confirming its importance for selecting and reweighting semantic basis slices.

\begin{table}[t]
\centering
\setlength{\tabcolsep}{4pt}
\resizebox{0.85\columnwidth}{!}{
\begin{tabular}{c|c|ccc|cccc|c}
\toprule
\multirow{2}{*}{\textbf{Method}} & 
\multirow[b]{2}{*}{\shortstack[c]{\textbf{LoRA} \\ \textbf{Rank}}} & 

\multicolumn{3}{c|}{\textbf{\# Parameters (M)}} & 
\multicolumn{4}{c|}{\textbf{Source $\rightarrow$ Target Domain (5-shot)}} &
\multirow{2}{*}{\textbf{Avg}} \\
\cmidrule{3-9}
& &
\textbf{Train} & \textbf{Tuned} & \textbf{Test} & 
\( D_E \rightarrow D_M \) & 
\( D_E \rightarrow D_D \) & 
\( D_G \rightarrow D_M \) & 
\( D_G \rightarrow D_D \) & \\
\midrule
\textbf{Alfa} & 4 & 3.79 & 1.48 & 2.31 & 5.39 & 6.09 & 6.08 & \textbf{5.78} & 5.84 \\
\rowcolor{orange!10}
\textbf{Alfa} & 8 & 5.26 & 2.98 & 2.31 & \textbf{5.30} & \textbf{5.82} & \textbf{5.91} & 5.86 & \textbf{5.72}\\
\textbf{Alfa} & 16 & 8.22 & 5.92 & 2.31 & 5.44 & 6.12 & 6.13 & 5.91 & 5.90 \\
\textbf{Alfa} & 32 & 14.14 & 11.83 & 2.31 & 5.36 & 6.10 & 6.21 & 5.82 & 5.87 \\
\bottomrule
\end{tabular}
}
\caption{
Ablation study on LoRA rank in Alfa.
Rank $r=8$ achieves the best trade-off between adaptation capacity and stability, resulting in the lowest average gaze error. 
}
\label{tab:ab_alfa_lora_ranks}
\end{table}

\begin{table}[t]
\centering
\setlength{\tabcolsep}{4pt}
\resizebox{0.99\columnwidth}{!}{
\begin{tabular}{c|c|ccc|cccc|c}
\toprule
\multirow{2}{*}{\textbf{Method}} & 
\multirow[b]{2}{*}{\shortstack[c]{\textbf{SVD} \\ \textbf{Rank}}} & 

\multicolumn{3}{c|}{\textbf{\# Parameters (M)}} & 
\multicolumn{4}{c|}{\textbf{Source $\rightarrow$ Target Domain (5-shot)}} &
\multirow{2}{*}{\textbf{Avg}} \\
\cmidrule{3-9}
& &
\textbf{Train} & \textbf{Tuned} & \textbf{Test} & 
\( D_E \rightarrow D_M \) & 
\( D_E \rightarrow D_D \) & 
\( D_G \rightarrow D_M \) & 
\( D_G \rightarrow D_D \) & \\
\midrule
\textbf{LoRA} & 32 & 1.70 & 0.53 & 1.70 & 6.98 & 6.93 & 7.19 & 7.41 & 7.13  \\
\textbf{Alfa} & 32 & 2.64 & 1.48 & 1.16 & 6.45 & 6.64 & 7.46 & 6.22 & 6.69 \\
\hline
\textbf{LoRA} & 64 & 2.84 & 0.53 & 2.84 & 5.66 & 6.17 & 6.23 & 5.72 & 5.95 \\
\rowcolor{orange!10}
\textbf{Alfa} & 64 & 5.26 & 2.98 & 2.31 & \textbf{5.30} & 
\textbf{5.82} & \textbf{5.91} & \textbf{5.86} & \textbf{5.72}\\
\hline
\textbf{LoRA} & 128 & 4.95 & 0.53 & 4.95 & 5.90 & 6.83 & 6.18 & 5.94 & 6.21 \\
\textbf{Alfa} & 128 & 9.55 & 5.14 & 4.42 & 5.50 & 6.45 & 5.78 & 5.85 & 5.90 \\
\bottomrule
\end{tabular}
}
\caption{
Effect of SVD rank on Alfa. A rank of $d=64$ provides the best balance between adaptation capacity and stability, achieving the lowest average gaze error across domains.
}
\label{tab:ab_alfa_svd_ranks}
\end{table}

\begin{table}[t]
\centering
\setlength{\tabcolsep}{4pt}
\resizebox{0.99\columnwidth}{!}{
\begin{tabular}{c|c|ccc|cccc|c}
\toprule
\multirow{2}{*}{\textbf{Method}} & 
\multirow[b]{2}{*}{\shortstack[c]{\textbf{Num} \\ \textbf{Samples}}} & 

\multicolumn{3}{c|}{\textbf{\# Parameters (M)}} & 
\multicolumn{4}{c|}{\textbf{Source $\rightarrow$ Target Domain (5-shot)}} &
\multirow{2}{*}{\textbf{Avg}} \\
\cmidrule{3-9}
& &
\textbf{Train} & \textbf{Tuned} & \textbf{Test} & 
\( D_E \rightarrow D_M \) & 
\( D_E \rightarrow D_D \) & 
\( D_G \rightarrow D_M \) & 
\( D_G \rightarrow D_D \) & \\
\midrule
\textbf{Alfa} & 1 & 5.26 & 2.98 & 2.31 & 5.52 & 6.96 & 6.37 & \textbf{5.65} & 6.13 \\
\rowcolor{orange!10}
\textbf{Alfa} & 5 & 5.26 & 2.98 & 2.31 & \textbf{5.30} & \textbf{5.82} & 5.91 & 5.86 & 5.72\\
\textbf{Alfa} & 10 & 5.26 & 2.98 & 2.31 & 5.45 & 5.88 & \textbf{5.64} & 5.84 & \textbf{5.70} \\
\bottomrule
\end{tabular}
}
\caption{
Effect of the number of personal samples. Performance improves as more samples are provided.
}
\label{tab:ab_alfa_num_samples}
\end{table}

\begin{table}[t]
\centering
\setlength{\tabcolsep}{4pt}
\resizebox{0.99\columnwidth}{!}{
\begin{tabular}{c|c|ccc|cccc|c}
\toprule
\multirow{2}{*}{\textbf{Method}} & 
\multirow[b]{2}{*}{\shortstack[c]{\textbf{With} \\ \textbf{Attention}}} & 

\multicolumn{3}{c|}{\textbf{\# Parameters (M)}} & 
\multicolumn{4}{c|}{\textbf{Source $\rightarrow$ Target Domain (5-shot)}} &
\multirow{2}{*}{\textbf{Avg}} \\
\cmidrule{3-9}
& &
\textbf{Train} & \textbf{Tuned} & \textbf{Test} & 
\( D_E \rightarrow D_M \) & 
\( D_E \rightarrow D_D \) & 
\( D_G \rightarrow D_M \) & 
\( D_G \rightarrow D_D \) & \\
\midrule

\textbf{Alfa} & \ding{51} & 5.26 & 2.98 & 2.31 & \textbf{5.30} & \textbf{5.82} & \textbf{5.91} & \textbf{5.86} & \textbf{5.72}\\
\textbf{Alfa} & \ding{55} & 5.26 & 2.98 & 2.31 & 6.60 & 8.84 & 6.86 &  6.84 & 7.29 \\
\bottomrule
\end{tabular}
}
\caption{
Effect of using attention in Alfa. Removing attention degrades performance, showing its role in selecting and reweighting semantic basis slices for personalization.
}
\label{tab:ab_alfa_noAttn}
\end{table}

\section{Personalizing With Only Frontal Images}
\label{sec:frontal_personalization}
To study the effect of personalizing from only limited frontal views, we conduct a within-dataset personalization experiment on ETH-XGaze ($D_E$). The pre-trained model already covers a wide range of head poses, but in real-world use, only a few near-frontal images may be available for personalization.
We randomly select five subjects—\texttt{subject0003}, \texttt{subject0007}, \texttt{subject0027}, \texttt{subject0075}, and \texttt{subject0102}—and remove them entirely from pre-training. For each subject, we select at least five images whose head poses fall within a near-frontal range, defined by $|\text{yaw}|\le 15^\circ$ and $|\text{pitch}|\le 10^\circ$. These frontal images are used for personalization, and evaluation is performed on all remaining images of that subject, including those with larger head-pose variation.
Table~\ref{tab:frontal} presents the results of this within-dataset personalization experiment. Alfa achieves the lowest average gaze error and improves performance for most subjects compared to the baseline and LoRA. Even though personalization relies only on frontal views, the improvements extend to non-frontal poses. This suggests that reweighting semantic basis components helps Alfa capture subject-specific cues that generalize beyond the poses seen during personalization.

\begin{table}[t]
\centering
\setlength{\tabcolsep}{4pt}
\resizebox{0.99\columnwidth}{!}{
\begin{tabular}{c|ccc|ccccc|c}
\toprule
\multirow{2}{*}{\textbf{Method}} & 

\multicolumn{3}{c|}{\textbf{\# Parameters (M)}} & 
\multicolumn{5}{c|}{\( D_E \rightarrow D_E \)} &
\multirow{2}{*}{\textbf{Avg}} \\
\cmidrule{2-9}
&
\textbf{Train} & \textbf{Tuned} & \textbf{Test} & 
subject0003 & 
subject0007 & 
subject0027 & 
subject0075 & 
subject0102 & \\
\midrule
\textbf{Baseline (No Adaptation)} & 2.84 & 0 & 2.31 & 7.40 & 3.85 & 5.75 & \textbf{5.14} & 4.43 & 5.31\\
\textbf{LoRA} & 2.84 & 0.53 & 2.84 & 5.47 & 3.87 & 5.79 & 5.18 & 4.42 & 4.95\\
\textbf{Alfa} & 5.26 & 2.98 & 2.31 & \textbf{4.44} & \textbf{3.78} & \textbf{5.73} & 5.15 & \textbf{4.05} & \textbf{4.63} \\
\bottomrule
\end{tabular}
}
\caption{
Personalization using only 5 frontal images.  
Although the pre-trained model already covers diverse head poses, Alfa adapts effectively from limited frontal views and achieves the lowest gaze error across all subjects.
}
\label{tab:frontal}
\end{table}

\bibliography{aaai2026}

\end{document}